\documentclass[letterpaper]{article} % DO NOT CHANGE THIS
\usepackage[submission]{aaai23}  % DO NOT CHANGE THIS
\usepackage{times}  % DO NOT CHANGE THIS
\usepackage{helvet}  % DO NOT CHANGE THIS
\usepackage{courier}  % DO NOT CHANGE THIS
\usepackage[hyphens]{url}  % DO NOT CHANGE THIS
\usepackage{graphicx} % DO NOT CHANGE THIS
\usepackage{natbib}  % DO NOT CHANGE THIS AND DO NOT ADD ANY OPTIONS TO IT
\usepackage{caption} % DO NOT CHANGE THIS AND DO NOT ADD ANY OPTIONS TO IT
\usepackage{algorithm}
\usepackage{algorithmic}

\usepackage{nicefrac}
\usepackage{microtype}
\usepackage{graphicx}
\usepackage{caption}
\usepackage{subcaption}
\usepackage{tikz-cd}
\usetikzlibrary{arrows}
\usepackage{comment}
\usepackage{amssymb}
\usepackage{tikz}
\usepackage{tikz-qtree,tikz-qtree-compat}
\usetikzlibrary{positioning}

\usepackage{amsmath}
\usepackage{amsthm}
\usepackage{tikz}
\usepackage{tikz-qtree,tikz-qtree-compat}

\makeatletter
\newtheorem*{rep@theorem}{\rep@title}
\newcommand{\newreptheorem}[2]{%
\newenvironment{rep#1}[1]{%
 \def\rep@title{#2 \ref{##1}}%
 \begin{rep@theorem}}%
 {\end{rep@theorem}}}
\makeatother

\newtheorem{theorem}{Theorem}
\newreptheorem{theorem}{Theorem}

\usetikzlibrary{positioning}
\newtheorem{definition}[theorem]{Definition}

\usepackage{stackengine}
\def\delequal{\mathrel{\ensurestackMath{\stackon[1pt]{=}{\scriptstyle\Delta}}}}

\usepackage{newfloat}
\usepackage{listings}
\DeclareCaptionStyle{ruled}{labelfont=normalfont,labelsep=colon,strut=off} % DO NOT CHANGE THIS
\floatstyle{ruled}
\newfloat{listing}{tb}{lst}{}
\floatname{listing}{Listing}
\title{Probabilities of Causation with Nonbinary Treatment and Effect}
\author{
    Ang Li,
    Judea Pearl
}
\affiliations{
    Cognitive Systems Laboratory, Department of Computer Science,\\
    University of California, Los Angeles,\\
    Los Angeles, California, USA.\\
    \{angli, judea\}@cs.ucla.edu
}

\usepackage{bibentry}
\begin{document}

\maketitle

\begin{abstract}
% This paper deals with the problem of estimating the probabilities of causation when treatment and effect are not binary. Tian and Pearl proposed sharp bounds for the probability of necessity and sufficiency (PNS), the probability of sufficiency (PS), and the probability of necessity (PN) using experimental and observational data. In this paper, we provide theoretical bounds for all types of probabilities of causation defined using the structural causal model (SCM), without restricting them to binary treatment and effect.
This paper deals with the problem of estimating the probabilities of causation when treatment and effect are not binary. Tian and Pearl derived sharp bounds for the probability of necessity and sufficiency (PNS), the probability of sufficiency (PS), and the probability of necessity (PN) using experimental and observational data. In this paper, we provide theoretical bounds for all types of probabilities of causation to multivalued treatments and effects. We further discuss examples where our bounds guide practical decisions and use simulation studies to evaluate how informative the bounds are for various combinations of data.
\end{abstract}

\section{Introduction}
In many areas of industry, marketing, and health science, the
probabilities of causation are widely used to solve decision-making problems. For example, Li and Pearl \cite{li:pea19-r488} proposed the “benefit function”, which is the payoff/cost associated with selecting an individual with given characteristics to identify a set of individuals who are most likely to exhibit a desired mode of behavior. In Li and Pearl's paper, the benefit function is a linear combination of the probabilities of causation with binary treatment and effect. For another example, Mueller and Pearl \cite{mueller:pea-r513} demonstrated that the probabilities of causation should be considered in personalized decision-making.

Consider the following motivating scenario: an elderly patient with cancer is faced with the choice of treatment to pursue. The options include surgery, chemotherapy, and radiation. The outcomes include ineffective, cured, and death. Given that the elderly patient has a high risk of death from cancer surgery, the patient wants to know the probability that he would be cured if he chose radiation, would die if he chose surgery, and nothing would change if he chose chemotherapy. Let $X$ denotes the treatment, where $x_1$ denotes surgery, $x_2$ denotes chemotherapy, and $x_3$ denotes radiation. Let $Y$ denotes the outcome, where $y_1$ denotes ineffective, $y_2$ denotes cured, and $y_3$ denotes death. The probability that the patient desires is the probability of causation, $P({y_3}_{x_1},{y_1}_{x_2},{y_2}_{x_3})$.

Pearl \cite{pearl1999probabilities} first defined three binary probabilities of causation (i.e., PNS, PN, and PS) using SCM \cite{galles1998axiomatic,halpern2000axiomatizing,pearl2009causality}. Tian and Pearl \cite{tian2000probabilities} then used observational and experimental data to bound those three probabilities of causation. Li and Pearl \cite{li:pea19-r488,li2022unit} provided formal proof of those bounds. Mueller, Li, and Pearl \cite{pearl:etal21-r505} recently proposed using covariate information and the causal structure to narrow the bounds of the probability of necessity and sufficiency. Dawid et al. \cite{dawid2017} also proposed using covariate information to narrow the bounds of the probability of necessity. 

All the above-mentioned studies are restricted to binary treatment and effect, limiting the application of probabilities of causation. Zhang, Tian, and Bareinboim \cite{zhang2022partial}, as well as Li and Pearl \cite{li2022bounds}, proposed nonlinear programming-based solutions to compute the bounds of nonbinary probabilities of causation numerically. However, the theoretical foundation of nonbinary probabilities of causation is still required, not only because numerical methods are limited by computational power but also because people are interested in the theoretical foundation due to further development and analysis. In this paper, we will introduce the theoretical bounds of any probabilities of causation defined using SCM without restricting them to binary treatment and effect.

\section{Preliminaries}
\label{related work}
In this section, we review the definitions for the three aspects of binary causation, as defined in \cite{pearl1999probabilities}. We use the language of counterfactuals in SCM, as defined in \cite{galles1998axiomatic,halpern2000axiomatizing}. 

We use $Y_x=y$ to denote the counterfactual sentence ``Variable $Y$ would have the value $y$, had $X$ been $x$". For the remainder of the paper, we use $y_x$ to denote the event $Y_x=y$, $y_{x'}$ to denote the event $Y_{x'}=y$, $y'_x$ to denote the event $Y_x=y'$, and $y'_{x'}$ to denote the event $Y_{x'}=y'$. We assume that experimental data will be summarized in the form of the causal effects such as $P(y_x)$ and observational data will be summarized in the form of the joint probability function such as $P(x,y)$. If not specified, the variable $X$ stands for treatment and the variable $Y$ stands for effect.

Three prominent probabilities of causation are as follows:
\begin{definition}[Probability of necessity (PN)]
Let $X$ and $Y$ be two binary variables in a causal model $M$, let $x$ and $y$ stand for the propositions $X=true$ and $Y=true$, respectively, and $x'$ and $y'$ for their complements. The probability of necessity is defined as the expression \cite{pearl1999probabilities}\\
\begin{eqnarray}
\text{PN} & \delequal & P(Y_{x'}=false|X=true,Y=true)\nonumber \\
& \delequal & P(y'_{x'}|x,y) \nonumber
\label{pn}
\end{eqnarray}
\end{definition}
% \par
% In other words, PN stands for the probability that event $y$ would not have occurred in the absence of event $x$, given that $x$ and $y$ did in fact occur.

% Note that lower case letters (e.g., $x,y$) stand for propositions (or events). PN has applications in epidemiology, legal reasoning, and artificial intelligence. Epidemiologists have long been concerned with estimating the probability that a certain case of disease is attributable to a particular exposure, which is normally interpreted counterfactually as ``the probability that disease would not have occurred in the absence of exposure, given that disease and exposure did in fact occur." This counterfactual notion is also used frequently in lawsuits, where legal responsibility is at the center of contention.
% \vspace{10pt}
\begin{definition}[Probability of sufficiency (PS)] \cite{pearl1999probabilities}
\begin{eqnarray}
\text{PS} \delequal P(y_x|y',x') \nonumber
\label{ps}
\end{eqnarray}
\end{definition}

% PS finds applications in policy analysis, artificial intelligence, and psychology.
% A policy maker may well be interested in the dangers that a certain exposure may present to the healthy population \cite{khoury1989measurement}. Counterfactually, this notion is expressed as the ``probability that a healthy unexposed individual would have gotten the disease had he/she been exposed." In psychology, PS serves as the basis for Cheng's \cite{cheng1997covariation} causal power theory \cite{glymour2013psychological}, which attempts to explain how humans judge causal strength among events. In artificial intelligence, PS plays a major role in the generation of explanations \cite{pearl2009causality}.
% \vspace{10pt}
\begin{definition}[Probability of necessity and sufficiency (PNS)] \cite{pearl1999probabilities}
\begin{eqnarray}
\text{PNS}\delequal P(y_x,y'_{x'}) \nonumber
\label{pns}
\end{eqnarray}
\end{definition}
\par
PNS stands for the probability that $y$ would respond to $x$ both ways, and therefore measures both the sufficiency and necessity of $x$ to produce $y$.

Tian and Pearl \cite{tian2000probabilities} provided tight bounds for PNS, PN, and PS using Balke's program \cite{balke1995probabilistic} (we will call them Tian-Pearl's bounds). Li and Pearl \cite{li:pea19-r488,li2022unit} provided theoretical proof of the tight bounds for PNS, PS, PN, and other binary probabilities of causation.

PNS, PN, and PS have the following tight bounds:\\

\begin{eqnarray*}
\max \left \{
\begin{array}{cc}
0, \\
P(y_x) - P(y_{x'}), \\
P(y) - P(y_{x'}), \\
P(y_x) - P(y)\\
\end{array}
\right \}
\le \text{PNS}
\label{pnslb}
\end{eqnarray*}

\begin{eqnarray*}
\text{PNS} \le \min \left \{
\begin{array}{cc}
 P(y_x), \\
 P(y'_{x'}), \\
P(x,y) + P(x',y'), \\
P(y_x) - P(y_{x'}), +\\
+ P(x, y') + P(x', y)
\end{array} 
\right \}
\label{pnsub}
\end{eqnarray*}

\begin{eqnarray*}
\max \left \{
\begin{array}{cc}
0, \\
\frac{P(y)-P(y_{x'})}{P(x,y)}
\end{array} 
\right \}
\le \text{PN}
\label{pnlb}
\end{eqnarray*}

\begin{eqnarray*}
\text{PN} \le
\min \left \{
\begin{array}{cc}
1, \\
\frac{P(y'_{x'})-P(x',y')}{P(x,y)} 
\end{array}
\right \}
\label{pnub}
\end{eqnarray*}

Note that we only consider PNS and PN here because the bounds of PS can easily be obtained by exchanging $x$ with $x'$ and $y$ with $y'$ in the bounds of PN. To obtain bounds for a specific population, defined by a set $C$ of characteristics, the expressions above should be modified by conditioning each term on $C=c$.

However, the above three probabilities of causation are unable to answer the query in our motivating example. In this paper, we demonstrate the bounds of any types of probabilities of causation. We illustrate the theorems by the order of the number of hypothetical terms (i.e., number of $y_x$ terms in the probability of causation). For example, the number of hypothetical terms in $P(y_x,{y'}_{x'})$ is $2$. The proof of all theorems is provided in the appendix.

\section{Probabilities of Causation with Single Hypothetical Term}
We start with four simple probabilities of causation with a single hypothetical term. Let $X$ denotes the treatment with potential values $x_1,...,x_m$ and $Y$ denotes the effect with potential values $y_1,...,y_n$. The four probabilities of causation with a single hypothetical term are $P({y_i}_{x_j}, y_i)$, $P({y_i}_{x_j}, y_k), s.t., i\ne k$, $P({y_i}_{x_j}, x_k),  s.t., j\ne k$, and $P({y_i}_{x_j}, y_k, x_m), s.t., m\ne j$. The following theorems define their bounds using observational and experimental data.

\begin{theorem}
\label{thm4}
Suppose variable $X$ has $m$ values $x_1,...,x_m$ and $Y$ has $n$ values $y_1,...,y_n$, then the probability of causation $P({y_i}_{x_j}, y_i)$, where $1 \le i \le n, 1 \le j \le m$, is bounded as following:
\begin{eqnarray*}
\max \left \{
\begin{array}{cc}
P(x_j, y_i), \\
P({y_i}_{x_j}) + P(y_i) - 1 \\
\end{array}
\right \}
\le P({y_i}_{x_j}, y_i)
\label{}
\end{eqnarray*}
\begin{eqnarray*}
P({y_i}_{x_j}, y_i) \le \min \left \{
\begin{array}{cc}
 P({y_i}_{x_j}), \\
 P(y_i) \\
\end{array} 
\right \}
\label{}
\end{eqnarray*}
\end{theorem}

\begin{theorem}
\label{thm5}
Suppose variable $X$ has $m$ values $x_1,...,x_m$ and $Y$ has $n$ values $y_1,...,y_n$, then the probability of causation $P({y_i}_{x_j}, y_k)$, where $1 \le i,k \le n, 1 \le j \le m, i\ne k$, is bounded as following:
\begin{eqnarray*}
\max \left \{
\begin{array}{cc}
0, \\
P({y_i}_{x_j}) + P(y_k) - 1, \\
\sum_{1\le p\le m,p\ne j}\max \left \{
\begin{array}{cc}
0, \\
P({y_i}_{x_j})\\
+ P(x_p,y_k) \\
- 1 + P(x_j)\\
- P(x_j,y_i) \\
\end{array}
\right \}
\end{array}
\right \}\nonumber\\
\le P({y_i}_{x_j}, y_k)
\label{}
\end{eqnarray*}
\begin{eqnarray*}
P({y_i}_{x_j}, y_k) \le \min \left \{
\begin{array}{cc}
 P({y_i}_{x_j}) - P(x_j, y_i), \\
 P(y_k) - P(y_k, x_j) \\
\end{array} 
\right \}
\label{}
\end{eqnarray*}
\end{theorem}

\begin{theorem}
\label{thm6}
Suppose variable $X$ has $m$ values $x_1,...,x_m$ and $Y$ has $n$ values $y_1,...,y_n$, then the probability of causation $P({y_i}_{x_j}, x_k)$, where $1 \le i \le n, 1 \le j,k \le m, j\ne k$, is bounded as following:
\begin{eqnarray*}
\max \left \{
\begin{array}{cc}
0, \\
P({y_i}_{x_j}) - P(x_j, y_i)\\
- 1 + P(x_j) + P(x_k) \\
\end{array}
\right \}
\le P({y_i}_{x_j}, x_k)
\label{}
\end{eqnarray*}
\begin{eqnarray*}
P({y_i}_{x_j}, x_k) \le \min \left \{
\begin{array}{cc}
 P({y_i}_{x_j}) - P(x_j, y_i), \\
 P(x_k) \\
\end{array} 
\right \}
\label{}
\end{eqnarray*}
\end{theorem}

\begin{theorem}
\label{thm7}
Suppose variable $X$ has $m$ values $x_1,...,x_m$ and $Y$ has $n$ values $y_1,...,y_n$, then the probability of causation $P({y_i}_{x_j}, y_k, x_p)$, where $1 \le i,k \le n, 1 \le j,p \le m, j\ne p$, is bounded as following:
\begin{eqnarray*}
\max \left \{
\begin{array}{cc}
0, \\
 P({y_i}_{x_j}) + P(x_p, y_k)\\
- 1 + P(x_j) - P(x_j, y_i) \\
\end{array}
\right \}
\le P({y_i}_{x_j}, y_k, x_p)
\label{}
\end{eqnarray*}
\begin{eqnarray*}
P({y_i}_{x_j}, y_k, x_p) \le \min \left \{
\begin{array}{cc}
 P({y_i}_{x_j}) - P(x_j, y_i), \\
 P(x_p, y_k) \\
\end{array} 
\right \}
\label{}
\end{eqnarray*}
\end{theorem}

Note that, there is no theorem for the probability of causation $P({y_i}_{x_j}, x_j)$ because $P({y_i}_{x_j}, x_j)$ simply equals $P(y_i, x_j)$. Moreover, Theorem \ref{thm7} is the general form of PS and PN. Besides, we do not have any theorem about conditional probabilities, because conditioning on observations does not change the properties of the bounds. For example, $P({y_i}_{x_j}| y_k, x_p)=P({y_i}_{x_j}, y_k, x_p)/P(y_k,x_p)$.

\section{Probabilities of Causation with Multi Hypothetical Terms}
In this section, we deal with four complicated probabilities of causation with multi hypothetical terms. They are $P({y_{i_1}}_{x_{j_1}},...,{y_{i_k}}_{x_{j_k}})$, $P({y_{i_1}}_{x_{j_1}},...,{y_{i_k}}_{x_{j_k}}, x_p)$, $P({y_{i_1}}_{x_{j_1}},...,{y_{i_k}}_{x_{j_k}}, y_q)$, and $P({y_{i_1}}_{x_{j_1}},...,{y_{i_k}}_{x_{j_k}}, x_p, y_q)$, s.t.,$j_1\ne ...\ne j_k \ne p$. Unlike the bounds in single hypothetical term cases, the bounds in this section are bounded recursively with cases with a smaller number of hypothetical terms. The following theorems provide the bounds using observational and experimental data.
\begin{theorem}
\label{thm8}
Suppose variable $X$ has $m$ values $x_1,...,x_m$ and $Y$ has $n$ values $y_1,...,y_n$, then the probability of causation $P({y_{i_1}}_{x_{j_1}},...,{y_{i_k}}_{x_{j_k}})$, where $1 \le i_1,...,i_k \le n, 1 \le j_1,...,j_k \le m, j_1\ne ... \ne j_k$, is bounded as following:
\begin{eqnarray*}
\max \left \{
\begin{array}{cc}
0, \\
\\
\sum_{1\le t\le k}P({y_{i_t}}_{x_{j_t}}) - k + 1, \\
\\
\max_{1\le t \le k} (LB(P({y_{i_1}}_{x_{j_1}},...,{y_{i_{t-1}}}_{x_{j_{t-1}}},\\
{y_{i_{t+1}}}_{x_{j_{t+1}}},...,{y_{i_k}}_{x_{j_k}}))\\
+ P({y_{i_t}}_{x_{i_t}}) - 1),\\
\\
\sum_{1\le p\le m, s.t., \exists r, 1\le r\le k, p=j_r }\\ LB(P({y_{i_1}}_{x_{j_1}},...,{y_{i_{r-1}}}_{x_{j_{r-1}}},\\
{y_{i_{r+1}}}_{x_{j_{r+1}}},...,{y_{i_k}}_{x_{j_k}}, x_{j_r}, y_{i_r})) + \\
\sum_{1\le p\le m, s.t., p \ne j_1 \ne ...\ne j_k}\\ LB(P({y_{i_1}}_{x_{j_1}},...,{y_{i_k}}_{x_{j_k}}, x_p))\\
\end{array}
\right \}\nonumber\\
\le P({y_{i_1}}_{x_{j_1}},...,{y_{i_k}}_{x_{j_k}})
\label{}
\end{eqnarray*}
\begin{eqnarray*}
P({y_{i_1}}_{x_{j_1}},...,{y_{i_k}}_{x_{j_k}}) \le \nonumber\\
\min \left \{
\begin{array}{cc}
 \min_{1\le t\le k} P({y_{i_t}}_{x_{j_t}}), \\
 \\
 \min_{1\le t \le k} UB(P({y_{i_1}}_{x_{j_1}},...,{y_{i_{t-1}}}_{x_{j_{t-1}}},\\
{y_{i_{t+1}}}_{x_{j_{t+1}}},...,{y_{i_k}}_{x_{j_k}})),\\
 \\
\sum_{1\le p\le m, s.t., \exists r, 1\le r\le k, p=j_r}\\ UB(P({y_{i_1}}_{x_{j_1}},...,{y_{i_{r-1}}}_{x_{j_{r-1}}},\\
{y_{i_{r+1}}}_{x_{j_{r+1}}},...,{y_{i_k}}_{x_{j_k}}, x_{j_r}, y_{i_r})) + \\
\sum_{1\le p\le m, s.t., p \ne j_1 \ne ...\ne j_k}\\ UB(P({y_{i_1}}_{x_{j_1}},...,{y_{i_k}}_{x_{j_k}}, x_p))\\
\end{array} 
\right \}
\label{}
\end{eqnarray*}
where,\\
LB$(f)$ denotes the lower bound of a function $f$ and UB$(f)$ denotes the upper bound of a function $f$. The bounds of $P({y_{i_1}}_{x_{j_1}},...,{y_{i_{r-1}}}_{x_{j_{r-1}}},{y_{i_{r+1}}}_{x_{j_{r+1}}},...,{y_{i_k}}_{x_{j_k}}, x_{j_r}, y_{j_r})$ are given by Theorem \ref{thm7} or \ref{thm11}, the bounds of $P({y_{i_1}}_{x_{j_1}},...,{y_{i_k}}_{x_{j_k}}, x_p)$ are given by Theorem \ref{thm6} or \ref{thm9}, and the bounds of $P({y_{i_1}}_{x_{j_1}},...,{y_{i_{t-1}}}_{x_{j_{t-1}}},{y_{i_{t+1}}}_{x_{j_{t+1}}},...,{y_{i_k}}_{x_{j_k}})$ are given by Theorem \ref{thm8} or experimental data if $k=2$. 
\end{theorem}

\begin{theorem}
\label{thm9}
Suppose variable $X$ has $m$ values $x_1,...,x_m$ and $Y$ has $n$ values $y_1,...,y_n$, then the probability of causation $P({y_{i_1}}_{x_{j_1}},...,{y_{i_k}}_{x_{j_k}},x_p)$, where $1 \le i_1,...,i_k \le n, 1 \le j_1,...,j_k,p \le m, j_1\ne ... \ne j_k \ne p$, is bounded as following:
\begin{eqnarray*}
\max \left \{
\begin{array}{cc}
0, \\
\\
\sum_{1\le t\le k}P({y_{i_t}}_{x_{j_t}}) +P(x_p) - k, \\
\\
\max_{1\le t \le k} (LB(P({y_{i_1}}_{x_{j_1}},...,{y_{i_{t-1}}}_{x_{j_{t-1}}},\\
{y_{i_{t+1}}}_{x_{j_{t+1}}},...,{y_{i_k}}_{x_{j_k}}))\\
+ LB(P({y_{i_t}}_{x_{i_t}},x_p)) - 1)\\
\end{array}
\right \}\nonumber\\
\le P({y_{i_1}}_{x_{j_1}},...,{y_{i_k}}_{x_{j_k}},x_p)
\label{}
\end{eqnarray*}
\begin{eqnarray*}
P({y_{i_1}}_{x_{j_1}},...,{y_{i_k}}_{x_{j_k}},x_p) \le \nonumber\\
\min \left \{
\begin{array}{cc}
 \min_{1\le t\le k} P({y_{i_t}}_{x_{j_t}}), \\
 \\
 P(x_p),\\
 \\
\min_{1\le t \le k} UB(P({y_{i_1}}_{x_{j_1}},...,{y_{i_{t-1}}}_{x_{j_{t-1}}},\\
{y_{i_{t+1}}}_{x_{j_{t+1}}},...,{y_{i_k}}_{x_{j_k}})),\\
\\
\min_{1\le t \le k} UB(P({y_{i_t}}_{x_{i_t}},x_p))\\
\end{array} 
\right \}
\label{}
\end{eqnarray*}
where,\\
LB$(f)$ denotes the lower bound of a function $f$ and UB$(f)$ denotes the upper bound of a function $f$. The bounds of $P({y_{i_1}}_{x_{j_1}},...,{y_{i_{t-1}}}_{x_{j_{t-1}}},{y_{i_{t+1}}}_{x_{j_{t+1}}},...,{y_{i_k}}_{x_{j_k}})$ are given by Theorem \ref{thm8} or experimental data if $k=2$ and the bounds of $P({y_{i_t}}_{x_{i_t}},x_p)$ are given by Theorem \ref{thm6}. 
\end{theorem}

\begin{theorem}
\label{thm10}
Suppose variable $X$ has $m$ values $x_1,...,x_m$ and $Y$ has $n$ values $y_1,...,y_n$, then the probability of causation $P({y_{i_1}}_{x_{j_1}},...,{y_{i_k}}_{x_{j_k}},y_q)$, where $1 \le i_1,...,i_k,q \le n, 1 \le j_1,...,j_k \le m, j_1\ne ... \ne j_k$, is bounded as following:
\begin{eqnarray*}
\max \left \{
\begin{array}{cc}
0,\\
\\
\sum_{1\le t\le k}P({y_{i_t}}_{x_{j_t}}) +P(y_q) - k, \\
\\
\max_{1\le t \le k} (LB(P({y_{i_1}}_{x_{j_1}},...,{y_{i_{t-1}}}_{x_{j_{t-1}}},\\
{y_{i_{t+1}}}_{x_{j_{t+1}}},...,{y_{i_k}}_{x_{j_k}}))\\
+ LB(P({y_{i_t}}_{x_{i_t}},y_q)) - 1),\\
\\
\sum_{1\le p\le m, \exists r, 1\le r\le k, p=j_r, q=i_r}\\ LB(P({y_{i_1}}_{x_{j_1}},...,{y_{i_{r-1}}}_{x_{j_{r-1}}},\\
{y_{i_{r+1}}}_{x_{j_{r+1}}},...,{y_{i_k}}_{x_{j_k}}, x_{j_r}, y_{i_r})) + \\
\sum_{1\le p\le m, s.t., p \ne j_1 \ne ...\ne j_k}\\ LB(P({y_{i_1}}_{x_{j_1}},...,{y_{i_k}}_{x_{j_k}}, x_p, y_q))\\
\end{array}
\right \}\nonumber\\
\le P({y_{i_1}}_{x_{j_1}},...,{y_{i_k}}_{x_{j_k}}, y_q)
\label{}
\end{eqnarray*}
\begin{eqnarray*}
P({y_{i_1}}_{x_{j_1}},...,{y_{i_k}}_{x_{j_k}},y_q) \le \nonumber\\
\min \left \{
\begin{array}{cc}
 \min_{1\le t\le k} P({y_{i_t}}_{x_{j_t}}), \\
 \\
 P(y_q),\\
 \\
\min_{1\le t \le k} UB(P({y_{i_1}}_{x_{j_1}},...,{y_{i_{t-1}}}_{x_{j_{t-1}}},\\
{y_{i_{t+1}}}_{x_{j_{t+1}}},...,{y_{i_k}}_{x_{j_k}})),\\
\\
\min_{1\le t \le k} UB(P({y_{i_t}}_{x_{i_t}},y_q)),\\
\\
\sum_{1\le p\le m, s.t., \exists r, 1\le r\le k, p=j_r, q=i_r}\\ UB(P({y_{i_1}}_{x_{j_1}},...,{y_{i_{r-1}}}_{x_{j_{r-1}}},\\
{y_{i_{r+1}}}_{x_{j_{r+1}}},...,{y_{i_k}}_{x_{j_k}}, x_{j_r}, y_{i_r})) + \\
\sum_{1\le p\le m, s.t., p \ne j_1 \ne ...\ne j_k}\\ UB(P({y_{i_1}}_{x_{j_1}},...,{y_{i_k}}_{x_{j_k}}, x_p, y_q))\\
\end{array} 
\right \}
\label{}
\end{eqnarray*}
where,\\
LB$(f)$ denotes the lower bound of a function $f$ and UB$(f)$ denotes the upper bound of a function $f$. The bounds of $P({y_{i_1}}_{x_{j_1}},...,{y_{i_{r-1}}}_{x_{j_{r-1}}},{y_{i_{r+1}}}_{x_{j_{r+1}}},...,{y_{i_k}}_{x_{j_k}}, x_{j_r}, y_{j_r})$, $P({y_{i_1}}_{x_{j_1}},...,{y_{i_k}}_{x_{j_k}}, x_p, y_q)$ are given by Theorem \ref{thm7} or \ref{thm11}, the bounds of $P({y_{i_1}}_{x_{j_1}},...,{y_{i_{t-1}}}_{x_{j_{t-1}}}, {y_{i_{t+1}}}_{x_{j_{t+1}}},...,{y_{i_k}}_{x_{j_k}})$ are given by Theorem \ref{thm8} or experimental data if $k=2$, and the bounds of $P({y_{i_t}}_{x_{i_t}},y_q)$ are given by Theorem \ref{thm4} or \ref{thm5}. 
\end{theorem}

\begin{theorem}
\label{thm11}
Suppose variable $X$ has $m$ values $x_1,...,x_m$ and $Y$ has $n$ values $y_1,...,y_n$, then the probability of causation $P({y_{i_1}}_{x_{j_1}},...,{y_{i_k}}_{x_{j_k}},x_p,y_q)$, where $1 \le i_1,...,i_k,q \le n, 1 \le j_1,...,j_k,p \le m, j_1\ne ... \ne j_k \ne p$, is bounded as following:
\begin{eqnarray*}
\max \left \{
\begin{array}{cc}
0, \\
\\
\sum_{1\le t\le k}P({y_{i_t}}_{x_{j_t}}) +P(x_p,y_q) - k, \\
\\
\max_{1\le t \le k} (LB(P({y_{i_1}}_{x_{j_1}},...,{y_{i_{t-1}}}_{x_{j_{t-1}}},\\
{y_{i_{t+1}}}_{x_{j_{t+1}}},...,{y_{i_k}}_{x_{j_k}}))\\
+ LB(P({y_{i_t}}_{x_{i_t}},x_p,y_q)) - 1)\\
\end{array}
\right \}\nonumber\\
\le P({y_{i_1}}_{x_{j_1}},...,{y_{i_k}}_{x_{j_k}},x_p,y_q)
\label{}
\end{eqnarray*}
\begin{eqnarray*}
P({y_{i_1}}_{x_{j_1}},...,{y_{i_k}}_{x_{j_k}},x_p,y_q) \le \nonumber\\
\min \left \{
\begin{array}{cc}
 \min_{1\le t\le k} P({y_{i_t}}_{x_{j_t}}), \\
 \\
 P(x_p,y_q),\\
 \\
\min_{1\le t \le k} UB(P({y_{i_1}}_{x_{j_1}},...,{y_{i_{t-1}}}_{x_{j_{t-1}}},\\
{y_{i_{t+1}}}_{x_{j_{t+1}}},...,{y_{i_k}}_{x_{j_k}})),\\
\\
\min_{1\le t \le k} UB(P({y_{i_t}}_{x_{i_t}},x_p,y_q))\\
\end{array} 
\right \}
\label{}
\end{eqnarray*}
where,\\
LB$(f)$ denotes the lower bound of a function $f$ and UB$(f)$ denotes the upper bound of a function $f$. The bounds of $P({y_{i_1}}_{x_{j_1}},...,{y_{i_{t-1}}}_{x_{j_{t-1}}}, {y_{i_{t+1}}}_{x_{j_{t+1}}},...,{y_{i_k}}_{x_{j_k}})$ are given by Theorem \ref{thm8} or experimental data if $k=2$ and the bounds of $P({y_{i_t}}_{x_{i_t}},x_p,y_q)$ are given by Theorem \ref{thm7}. 
\end{theorem}

Note that each theorem contains nonrecursively and recursively parts. The nonrecursively parts directly follow the Frechet inequalities. If it is realized that the nonrecursive parts are sufficient for decision-making, the recursive parts can be ignored. Furthermore, the recursive parts in theorems are guaranteed to reduce the number of hypothetical terms in the probabilities of causation by $1$; therefore, the recursive parts can reach the single hypothetical term cases in Theorems \ref{thm4} to \ref{thm7}. Moreover, Theorem \ref{thm8} is the general form of PNS.

\section{Examples}
In this section, we show how the presented theorems can be used in applications. We start with our motivating example.
\subsection{Choice of Treatment}
An elderly patient with cancer is faced with the choice of treatment. The options from the hospital include surgery, chemotherapy, and radiation. The outcomes include ineffective, cured, and death. Given the elder patient's high risk of death from cancer surgery, the doctor of the hospital suggested radiation to the patient. So, the patient wants to know the probability that he would be cured if he chose radiation, that would die if he chose surgery, and that nothing would change if he chose chemotherapy. 

Let $X$ denotes the treatment, where $x_1$ denotes surgery, $x_2$ denotes chemotherapy, and $x_3$ denotes radiation. Let $Y$ denotes the outcome, where $y_1$ denotes ineffective, $y_2$ denotes cured, and $y_3$ denotes death. The probability that the patient desires is the probability of causation, $P({y_3}_{x_1},{y_1}_{x_2},{y_2}_{x_3})$.

The doctor provided an experimental study of $900$ elderly patients where all the patients were forced to take treatment. The results are shown in Table \ref{tb1}.

\begin{table}
\centering
\begin{tabular}{|c|c|c|c|}
\hline 
&Surgery&Chemotherapy&Radiation\\
\hline
Ineffective&\begin{tabular}{c}$80$\\Patients\end{tabular}&\begin{tabular}{c}$184$\\Patients\end{tabular}&\begin{tabular}{c}$87$\\Patients\end{tabular}\\
\hline
Cured&\begin{tabular}{c}$7$\\Patients\end{tabular}&\begin{tabular}{c}$29$\\Patients\end{tabular}&\begin{tabular}{c}$189$\\Patients\end{tabular}\\
\hline
Death&\begin{tabular}{c}$213$\\Patients\end{tabular}&\begin{tabular}{c}$87$\\Patients\end{tabular}&\begin{tabular}{c}$24$\\Patients\end{tabular}\\
\hline
Overall&\begin{tabular}{c}$300$\\Patients\end{tabular}&\begin{tabular}{c}$300$\\Patients\end{tabular}&\begin{tabular}{c}$300$\\Patients\end{tabular}\\
\hline
\end{tabular}
\caption{Experimental data collected by the hospital. Here, $300$ patients were forced to receive surgery, $300$ patients were forced to receive chemotherapy, and $300$ patients were forced to receive radiation.}
\label{tb1}
\end{table}

The doctor also provided an observational study of $900$ elderly patients, where all the patients were open to all treatments and chose the treatment by themselves. The results are shown in Table \ref{tb2}.

\begin{table}
\centering
\begin{tabular}{|c|c|c|c|}
\hline 
&Surgery&Chemotherapy&Radiation\\
\hline
Ineffective&\begin{tabular}{c}$238$\\Patients\end{tabular}&\begin{tabular}{c}$10$\\Patients\end{tabular}&\begin{tabular}{c}$147$\\Patients\end{tabular}\\
\hline
Cured&\begin{tabular}{c}$20$\\Patients\end{tabular}&\begin{tabular}{c}$77$\\Patients\end{tabular}&\begin{tabular}{c}$72$\\Patients\end{tabular}\\
\hline
Death&\begin{tabular}{c}$7$\\Patients\end{tabular}&\begin{tabular}{c}$259$\\Patients\end{tabular}&\begin{tabular}{c}$70$\\Patients\end{tabular}\\
\hline
Overall&\begin{tabular}{c}$265$\\Patients\end{tabular}&\begin{tabular}{c}$346$\\Patients\end{tabular}&\begin{tabular}{c}$289$\\Patients\end{tabular}\\
\hline
\end{tabular}
\caption{Observational data collected by the hospital. Here, $900$ patients were free to choose one of the three treatments by themselves; $265$ patients chose surgery, $346$ patients chose chemotherapy, and $289$ patients chose radiation.}
\label{tb2}
\end{table}

The experimental data provide the following estimates:\\
\begin{eqnarray*}
P({y_1}_{x_1}) = 80 / 300,P({y_2}_{x_1}) = 7 / 300,\\
P({y_3}_{x_1}) = 213 / 300,P({y_1}_{x_2}) = 184 / 300,\\
P({y_2}_{x_2}) = 29 / 300,P({y_3}_{x_2}) = 87 / 300,\\
P({y_1}_{x_3}) = 87 / 300,P({y_2}_{x_3}) = 189 / 300,\\
P({y_3}_{x_3}) = 24 / 300.
\label{}
\end{eqnarray*}

Here, all three experimental estimates, $P({y_3}_{x_1})$, $P({y_1}_{x_2})$, and $P({y_2}_{x_3}))$, in the target probability of causation are higher than $0.5$, which may give us the sense that the target probability of causation, $P({y_3}_{x_1},{y_1}_{x_2},{y_2}_{x_3})$, would be high.

The observational data provide the following estimates:\\
\begin{eqnarray*}
P(x_1,y_1) = 238 / 900,P(x_1,y_2) = 20 / 900,\\
P(x_1,y_3) = 7 / 900,P(x_2,y_1) = 10 / 900,\\
P(x_2,y_2) = 77 / 900,P(x_2,y_3) = 259 / 900,\\
P(x_3,y_1) = 147 / 900,P(x_3,y_2) = 72 / 900,\\
P(x_3,y_3) = 70 / 900.
\label{}
\end{eqnarray*}

We then plug the estimates into Theorem \ref{thm8} (see the appendix for the detailed calculations). We obtain the bounds of the target probability of causation as follows:
% Note that Theorem \ref{thm8} contains three parts, one non-recursive part, $[\max\{0, \sum_{1\le t\le k}P({y_{i_t}}_{x_{j_t}}) - k + 1,\},\min_{1\le t\le k} P({y_{i_t}}_{x_{j_t}})]$, and two recursive parts. The non-recursive part provides the bounds as following:
% \begin{eqnarray*}
% 0\le P({y_3}_{x_1},{y_1}_{x_2},{y_2}_{x_3}) \le 0.613
% \label{}
% \end{eqnarray*}

% The upper bound, $0.613$, is still hard for decision making, therefore, we continue to compute the bounds with two recursive parts (see appendix for detailed computation), which the first recursive part provides the bounds as following:
% \begin{eqnarray*}
% 0\le P({y_3}_{x_1},{y_1}_{x_2},{y_2}_{x_3}) \le 0.099
% \label{}
% \end{eqnarray*}

% and the second recursive part provides the boudns as following: 
% \begin{eqnarray*}
% 0\le P({y_3}_{x_1},{y_1}_{x_2},{y_2}_{x_3}) \le 0.340
% \label{}
% \end{eqnarray*}

% Thus, the final bounds should be: 
\begin{eqnarray*}
0\le P({y_3}_{x_1},{y_1}_{x_2},{y_2}_{x_3}) \le 0.099
\label{}
\end{eqnarray*}

In conclusion, the probability that the patient would be cured if he chose radiation, that would die if he chose surgery, and that nothing would change if he chose chemotherapy is below $0.099$, implying that the patient should not consider radiation as a treatment option.

\subsection{Change of Institute}
Bob is looking for a job on the job market. There are three institutes, say A, B, and C, that offer courses to help people prepare for job searches. Bob went to one of the institutes, A, and took the course, but he still failed on the job market. Thus, Bob wonders if these courses improve his chance of getting a job. What would happen if he chose the other two institutes?

Let $X$ denotes which institute a person is chosen, where $x_1$ denotes that no institute is chosen, $x_2$ denotes institute A, $x_3$ denotes institute B, and $x_4$ denotes institute C. Let $Y$ denotes whether a person gets a job, where $y_1$ denotes success in job seeking, and $y_2$ denotes failure in job seeking. Therefore, Bob's questions become the following two probabilities of causation, $P({y_1}_{x_3}|x_2, y_2)$ and $P({y_1}_{x_4}|x_2, y_2)$.

All institutes provided experimental and observational studies to illustrate their effectiveness. Bob summarized the studies in Tables \ref{tb3} and \ref{tb4}.

\begin{table}
\centering
\begin{tabular}{|c|c|c|c|}
\hline 
&Success&Failure&Overall\\
\hline
No institute&\begin{tabular}{c}$53$\\People\end{tabular}&\begin{tabular}{c}$247$\\People\end{tabular}&\begin{tabular}{c}$300$\\People\end{tabular}\\
\hline
Institute A&\begin{tabular}{c}$269$\\People\end{tabular}&\begin{tabular}{c}$31$\\People\end{tabular}&\begin{tabular}{c}$300$\\People\end{tabular}\\
\hline
Institute B&\begin{tabular}{c}$234$\\People\end{tabular}&\begin{tabular}{c}$66$\\People\end{tabular}&\begin{tabular}{c}$300$\\People\end{tabular}\\
\hline
Institute C&\begin{tabular}{c}$151$\\People\end{tabular}&\begin{tabular}{c}$149$\\People\end{tabular}&\begin{tabular}{c}$300$\\People\end{tabular}\\
\hline
\end{tabular}
\caption{Experimental data collected by Bob. Here, $300$ people were forced to take no course, $300$ people were forced to take a course at institute A, $300$ people were forced to take a course at institute B, and $300$ people were forced to take a course at institute C.}
\label{tb3}
\end{table}

The experimental data provide the estimates:\\
\begin{eqnarray*}
P({y_1}_{x_1}) = 53 / 300,P({y_2}_{x_1}) = 247 / 300,\\
P({y_1}_{x_2}) = 269 / 300,P({y_2}_{x_2}) = 31 / 300,\\
P({y_1}_{x_3}) = 234 / 300,P({y_2}_{x_3}) = 66 / 300,\\
P({y_1}_{x_4}) = 151 / 300,P({y_2}_{x_4}) = 149 / 300.
\label{}
\end{eqnarray*}

The observational data provide the estimates:\\
\begin{eqnarray*}
P(x_1,y_1) = 92 / 1200,P(x_1,y_2) = 58 / 1200,\\
P(x_2,y_1) = 55 / 1200,P(x_2,y_2) = 118 / 1200,\\
P(x_3,y_1) = 24 / 1200,P(x_3,y_2) = 231 / 1200,\\
P(x_4,y_1) = 599 / 1200,P(x_4,y_2) = 23 / 1200.
\label{}
\end{eqnarray*}

\begin{table}
\centering
\begin{tabular}{|c|c|c|c|}
\hline 
&Success&Failure&Overall\\
\hline
No institute&\begin{tabular}{c}$92$\\People\end{tabular}&\begin{tabular}{c}$58$\\People\end{tabular}&\begin{tabular}{c}$150$\\People\end{tabular}\\
\hline
Institute A&\begin{tabular}{c}$55$\\People\end{tabular}&\begin{tabular}{c}$118$\\People\end{tabular}&\begin{tabular}{c}$173$\\People\end{tabular}\\
\hline
Institute B&\begin{tabular}{c}$24$\\People\end{tabular}&\begin{tabular}{c}$231$\\People\end{tabular}&\begin{tabular}{c}$255$\\People\end{tabular}\\
\hline
Institute C&\begin{tabular}{c}$599$\\People\end{tabular}&\begin{tabular}{c}$23$\\People\end{tabular}&\begin{tabular}{c}$622$\\People\end{tabular}\\
\hline
\end{tabular}
\caption{Observational data collected by Bob. Here, $1200$ people were open to all institutes, $150$ people chose to take no course, $173$ people chose to take a course at institute A, $255$ people chose to take a course at institute B, and $622$ people chose to take a course at institute C.}
\label{tb4}
\end{table}

Based on the experimental study, institute A claims that taking their course increased the success rate of finding a job from $0.177$ to $0.897$ and institute B claims that taking their course increased the success rate of finding a job from $0.177$ to $0.780$. Based on the observational study, institute C claims that taking their course increased the success rate of finding a job from $0.613$ to $0.963$. All of these seem useful to the job seeker, which is why Bob chose institute A previously. However, he still failed in the job market.

Now, consider the following two probabilities of causation, $P({y_1}_{x_3}|x_2, y_2)=P({y_1}_{x_3},x_2, y_2)/P(x_2,y_2)$, \\ $P({y_1}_{x_4}|x_2, y_2)=P({y_1}_{x_4},x_2, y_2)/P(x_2,y_2)$,\\ What would be the probability of success if he had chosen the other two institutes?

We plug the experimental and observational estimates into Theorem \ref{thm7} to obtain the following bounds:
\begin{eqnarray*}
0.720 \le P({y_1}_{x_3}|x_2, y_2) \le 1,\\
0 \le P({y_1}_{x_4}|x_2, y_2) \le 0.042.\\
\label{}
\end{eqnarray*}

Now Bob can see why he should change the institute to B.

\subsection{Effectiveness of Vaccine}
A clinical study is conducted to test the effectiveness of the vaccine. The treatment includes vaccinated and unvaccinated. The outcomes include uninfected by the virus, asymptomatic infected, infected with mild symptoms, and infected in a severe condition.

The goal of the clinical study is to learn the probability that a patient would be infected in a severe condition if unvaccinated and would be uninfected if vaccinated, the probability that a patient would be infected in a severe condition if unvaccinated and would be asymptomatic infected if vaccinated, and the probability that a patient would be infected in a severe condition if unvaccinated and would be infected with mild symptoms if vaccinated.

Let $X$ denotes vaccination with $x_1$ being vaccinated and $x_2$ being unvaccinated and $Y$ denotes the outcome, where $y_1$ denotes uninfected by the virus, $y_2$ denotes asymptomatic infected, $y_3$ denotes infected with mild symptoms, and $y_4$ denotes infected in a severe condition. The probabilities of causation we want to evaluate are $P({y_1}_{x_1},{y_4}_{x_2})$, $P({y_2}_{x_1},{y_4}_{x_2})$, and $P({y_3}_{x_1},{y_4}_{x_2})$.

The experimental and observational data of the clinical study are summarized in Tables \ref{tb5} and \ref{tb6}, respectively.

\begin{table}
\centering
\begin{tabular}{|c|c|c|}
\hline 
&Vaccinated&Unvaccinated\\
\hline
Uninfected&\begin{tabular}{c}$205$\\People\end{tabular}&\begin{tabular}{c}$27$\\People\end{tabular}\\
\hline
Asymptomatic&\begin{tabular}{c}$46$\\People\end{tabular}&\begin{tabular}{c}$122$\\People\end{tabular}\\
\hline
Mild Symptoms&\begin{tabular}{c}$343$\\People\end{tabular}&\begin{tabular}{c}$87$\\People\end{tabular}\\
\hline
Severe Condition&\begin{tabular}{c}$6$\\People\end{tabular}&\begin{tabular}{c}$364$\\People\end{tabular}\\
\hline
Overall&\begin{tabular}{c}$600$\\People\end{tabular}&\begin{tabular}{c}$600$\\People\end{tabular}\\
\hline
\end{tabular}
\caption{Experimental data of the clinical study. Here, $600$ people were forced to take the vaccine and $600$ people were forced to take no vaccine.}
\label{tb5}
\end{table}

\begin{table}
\centering
\begin{tabular}{|c|c|c|}
\hline 
&Vaccinated&Unvaccinated\\
\hline
Uninfected&\begin{tabular}{c}$6$\\People\end{tabular}&\begin{tabular}{c}$52$\\People\end{tabular}\\
\hline
Asymptomatic&\begin{tabular}{c}$74$\\People\end{tabular}&\begin{tabular}{c}$243$\\People\end{tabular}\\
\hline
Mild Symptoms&\begin{tabular}{c}$632$\\People\end{tabular}&\begin{tabular}{c}$147$\\People\end{tabular}\\
\hline
Severe Condition&\begin{tabular}{c}$5$\\People\end{tabular}&\begin{tabular}{c}$41$\\People\end{tabular}\\
\hline
Overall&\begin{tabular}{c}$717$\\People\end{tabular}&\begin{tabular}{c}$483$\\People\end{tabular}\\
\hline
\end{tabular}
\caption{Observational data of the clinical study. Here, $1200$ people were free to the vaccine. $717$ people chose to take the vaccine and $483$ people chose to take no vaccine.}
\label{tb6}
\end{table}

Based on the clinical study, the researcher of the vaccine claimed that the vaccine is effective in controlling the severe condition, the number of patients with a severe condition dropped from $364$ to only $6$. Besides, some of the patients would be even uninfected because the number of uninfected people increased from $27$ to $205$.

Now, consider the probability that a patient would be in a severe condition if unvaccinated and would be uninfected by virus if vaccinated, $P({y_1}_{x_1},{y_4}_{x_2})$, the probability that a patient would be in a severe condition if unvaccinated and would be asymptomatic infected if vaccinated, $P({y_2}_{x_1},{y_4}_{x_2})$, and the probability that a patient would be in a severe condition if unvaccinated and would be infected with mild symptoms if vaccinated, $P({y_3}_{x_1},{y_4}_{x_2})$.

The experimental data provide the following estimates:\\
\begin{eqnarray*}
P({y_1}_{x_1}) = 205 / 600,P({y_2}_{x_1}) = 46 / 600,\\
P({y_3}_{x_1}) = 343 / 600,P({y_4}_{x_1}) = 6 / 600,\\
P({y_1}_{x_2}) = 27 / 600,P({y_2}_{x_2}) = 122 / 600,\\
P({y_3}_{x_2}) = 87 / 600,P({y_4}_{x_2}) = 364 / 600.
\label{}
\end{eqnarray*}

The observational data provide the following estimates:\\
\begin{eqnarray*}
P(x_1,y_1) = 6 / 1200,P(x_1,y_2) = 74 / 1200,\\
P(x_1,y_3) = 632 / 1200,P(x_1,y_4) = 5 / 1200,\\
P(x_2,y_1) = 52 / 1200,P(x_2,y_2) = 243 / 1200,\\
P(x_2,y_3) = 147 / 1200,P(x_2,y_4) = 41 / 1200.
\label{}
\end{eqnarray*}

We plug the estimates into Theorem \ref{thm8} to obtain the bounds:
\begin{eqnarray*}
0 \le P({y_1}_{x_1},{y_4}_{x_2}) \le 0.039\\
0.037 \le P({y_2}_{x_1},{y_4}_{x_2}) \le 0.077\\
0.502 \le P({y_3}_{x_1},{y_4}_{x_2}) \le 0.561.
\label{}
\end{eqnarray*}

Thus, the probability of causation that a patient would be in a severe condition if unvaccinated and would be uninfected if vaccinated is at most $0.039$, the probability that a patient would be in a severe condition if unvaccinated and would be asymptomatic infected if vaccinated is at most $0.077$, and the probability that a patient would be in a severe condition if unvaccinated and would be infected with mild symptoms if vaccinated is at least $0.502$.

We conclude that the vaccine is effective in controlling the severe condition, but can only make it infected with mild symptoms. The vaccine is ineffective for uninfected and asymptomatic infected if the patient would be in a severe condition if unvaccinated.

\section{Simulated Results}
\label{simres}
In this section, we show the quality of the proposed bounds of the probabilities of causation.

\begin{figure}
\centering
\includegraphics[width=0.499\textwidth]{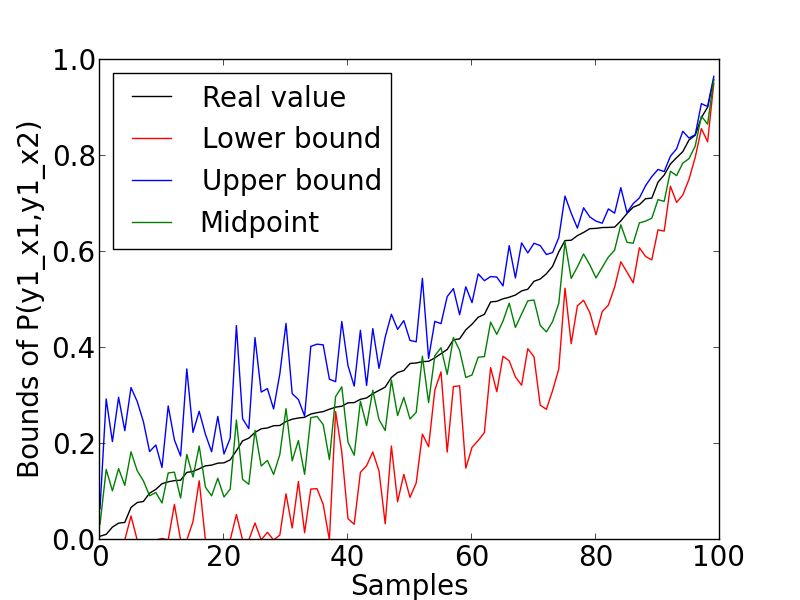}
\caption{Bounds of the $P({y_1}_{x_1},{y_1}_{x_2})$ for $100$ samples out of $1000$.}
\label{res1}
\end{figure}

We set $m=2$ (i.e., $X$ has two values) and $n=3$ (i.e., $Y$ has three values). We focus on the probability of causation, $P({y_1}_{x_1},{y_1}_{x_2})$. We randomly generated $1000$ samples of $P({y_1}_{x_1},{y_1}_{x_2})$. For each sample, we then generated sample distributions (observational data and experimental data) compatible with the $P({y_1}_{x_1},{y_1}_{x_2})$ (see the appendix for the generating algorithm). The advantage of this generating process is that we have the real value of the probability of causation for comparison. The generating algorithm ensures that the experimental data and observational data satisfy the general relation (i.e., $P(x,y|c)\le P(y_x|c) \le P(x,y|c) + 1 - P(x|c)$). For a sample $i$, let $[a_i,b_i]$ be the bounds of the $P({y_1}_{x_1},{y_1}_{x_2})$ obtained from the proposed theorems. We summarized the following criteria for each sample as illustrated in Figure \ref{res1}:
\begin{itemize}
    \item lower bound : $a_i$;
    \item upper bound : $b_i$;
    \item midpoint : $(a_i+b_i)/2$;
    \item real value;
\end{itemize}

From the Figure \ref{res1}, it is clear that the proposed bounds are a good estimation of the real probability of causation. The lower and upper bounds are closely around the real value and the midpoints are almost identified with the real value. Besides, the average gap of the bounds, $\frac{\sum(b_i-a_i)}{1000}$, is $0.228$, which make the bounds convincing.

\section{Discussion}
We demonstrated that nonbinary probabilities of causation help decision-maker in applications. However, we must discuss some properties of our proposed theorems further.

First, Tian-Pearl's bounds for PNS, PN, and PS are tight, implying that the bounds cannot be beaten if no additional assumption is made. However, the proposed theorems are not tight bounds, except for Theorem \ref{thm7} (Theorem \ref{thm7} can yield the bounds of PNS, PN, and PS, if Theorem \ref{thm7} is not tight, then the bounds of PNS, PN, and PS are not tight). The main contribution of this paper is to first provide theoretical bounds for nonbinary probabilities of causation. Researchers and decision-makers require theoretical bounds. We are happy that researchers can improve or prove the tightness of our bounds in the future.

Second, Theorems \ref{thm8} to \ref{thm11} contain recursive bounds. One may be concerned about the computation complexity. However, suppose the number of hypothetical terms is $k$, then the maximum number of  probabilities of causation considered in the recursion is $2^{(k+2)}$, where $k$ is usually small.

\section{Conclusion}
We demonstrated how to obtain bounds for any probabilities of causation defined using SCM with nonbinary treatment and effect. We derived eight theorems to deliver reasonable bounds. Both examples and simulated studies are provided to support the proposed theorems.

\section{Acknowledgements}
This research was supported in parts by grants from the National Science
Foundation [\#IIS-2106908], Office of Naval Research [\#N00014-17-S-12091
and \#N00014-21-1-2351], and Toyota Research Institute of North America
[\#PO-000897].

% One future direction of this work could be the statistical property of the proposed bounds. How tight in general the bounds would be? Does it enough to make decisions? Which data, experimental or observational, would affect the bounds more? Does the number of hypothetical terms in the probabilities of causation would affect the quality of the bounds?

% Another future direction could be to improve the bounds using covariate information similar to Mueller, Li, and Pearl \cite{pearl:etal21-r505} did to PNS, PN, and PS.

% Use \bibliography{yourbibfile} instead or the References section will not appear in your paper
%\bibliographystyle{named}
\bibliography{aaai23.bib}

\begin{thebibliography}{13}
\providecommand{\natexlab}[1]{#1}

\bibitem[{Balke(1995)}]{balke1995probabilistic}
Balke, A.~A. 1995.
\newblock \emph{Probabilistic counterfactuals: semantics, computation, and
  applications}.
\newblock University of California, Los Angeles.

\bibitem[{Dawid, Musio, and Murtas(2017)}]{dawid2017}
Dawid, P.; Musio, M.; and Murtas, R. 2017.
\newblock The Probability of Causation.
\newblock \emph{Law, Probability and Risk}, (16): 163--179.

\bibitem[{Galles and Pearl(1998)}]{galles1998axiomatic}
Galles, D.; and Pearl, J. 1998.
\newblock An axiomatic characterization of causal counterfactuals.
\newblock \emph{Foundations of Science}, 3(1): 151--182.

\bibitem[{Halpern(2000)}]{halpern2000axiomatizing}
Halpern, J.~Y. 2000.
\newblock Axiomatizing causal reasoning.
\newblock \emph{Journal of Artificial Intelligence Research}, 12: 317--337.

\bibitem[{Li and Pearl(2019)}]{li:pea19-r488}
Li, A.; and Pearl, J. 2019.
\newblock Unit Selection Based on Counterfactual Logic.
\newblock In \emph{Proceedings of the Twenty-Eighth International Joint
  Conference on Artificial Intelligence, {IJCAI-19}}, 1793--1799. International
  Joint Conferences on Artificial Intelligence Organization.

\bibitem[{Li and Pearl(2022{\natexlab{a}})}]{li2022bounds}
Li, A.; and Pearl, J. 2022{\natexlab{a}}.
\newblock Bounds on causal effects and application to high dimensional data.
\newblock In \emph{Proceedings of the AAAI Conference on Artificial
  Intelligence}, volume~36, 5773--5780.

\bibitem[{Li and Pearl(2022{\natexlab{b}})}]{li2022unit}
Li, A.; and Pearl, J. 2022{\natexlab{b}}.
\newblock Unit selection with causal diagram.
\newblock In \emph{Proceedings of the AAAI Conference on Artificial
  Intelligence}, volume~36, 5765--5772.

\bibitem[{Mueller and Pearl(2022)}]{mueller:pea-r513}
Mueller; and Pearl. 2022.
\newblock Personalized Decision Making -- A Conceptual Introduction.
\newblock Technical Report R-513, Department of Computer Science, University of
  California, Los Angeles, CA.

\bibitem[{Mueller, Li, and Pearl(2021)}]{pearl:etal21-r505}
Mueller, S.; Li, A.; and Pearl, J. 2021.
\newblock Causes of effects: Learning individual responses from population
  data.
\newblock Technical Report R-505,
  {$<$http://ftp.cs.ucla.edu/pub/stat\_ser/r505.pdf$>$}, Department of Computer
  Science, University of California, Los Angeles, CA.
\newblock Forthcoming, Proceedings of IJCAI-2022.

\bibitem[{Pearl(1999)}]{pearl1999probabilities}
Pearl, J. 1999.
\newblock Probabilities of Causation: Three Counterfactual Interpretations and
  Their Identification.
\newblock \emph{Synthese}, 93--149.

\bibitem[{Pearl(2009)}]{pearl2009causality}
Pearl, J. 2009.
\newblock \emph{Causality}.
\newblock Cambridge university press, 2nd edition.

\bibitem[{Tian and Pearl(2000)}]{tian2000probabilities}
Tian, J.; and Pearl, J. 2000.
\newblock Probabilities of causation: Bounds and identification.
\newblock \emph{Annals of Mathematics and Artificial Intelligence}, 28(1-4):
  287--313.

\bibitem[{Zhang, Tian, and Bareinboim(2022)}]{zhang2022partial}
Zhang, J.; Tian, J.; and Bareinboim, E. 2022.
\newblock Partial counterfactual identification from observational and
  experimental data.
\newblock In \emph{International Conference on Machine Learning}, 26548--26558.
  PMLR.

\end{thebibliography}
%\nobibliography{aaai23}

%\section{Acknowledgments}
\clearpage
\newpage
\appendix
\section{Appendix}
\subsection{Proof of Theorems}

\begin{reptheorem}{thm4}
Suppose variable $X$ has $m$ values $x_1,...,x_m$ and $Y$ has $n$ values $y_1,...,y_n$, then the probability of causation $P({y_i}_{x_j}, y_i)$, where $1 \le i \le n, 1 \le j \le m$, is bounded as following:
\begin{eqnarray}
\max \left \{
\begin{array}{cc}
P(x_j, y_i), \\
P({y_i}_{x_j}) + P(y_i) - 1 \\
\end{array}
\right \}
\le P({y_i}_{x_j}, y_i)
\label{t4e1}
\end{eqnarray}
\begin{eqnarray}
P({y_i}_{x_j}, y_i) \le \min \left \{
\begin{array}{cc}
 P({y_i}_{x_j}), \\
 P(y_i) \\
\end{array} 
\right \}
\label{t4e2}
\end{eqnarray}
\begin{proof}
By Fréchet Inequalities, we have,
\begin{eqnarray*}
 &&P(A, B) \ge \max\{0, P(A)+P(B)-1\},\\
 &&P(A, B)\le \min\{P(A),P(B)\}.
\end{eqnarray*}
Thus, we have,
\begin{eqnarray}
 &&P({y_i}_{x_j}, y_i) \ge \max\{0, P({y_i}_{x_j}) + P(y_i) - 1\}\label{t4e3},\\
 &&P({y_i}_{x_j}, y_i)\le \min\{P({y_i}_{x_j},P(y_i)\}.\nonumber
\end{eqnarray}
Therefore, Equation \ref{t4e2} holds.\\
We also have,
\begin{eqnarray*}
P({y_i}_{x_j}, y_i)&\ge&P({y_i}_{x_j}, y_i,x_j)\\
&=&P(x_j,y_i)\\
&\ge&0.
\end{eqnarray*}
Combine with Equation \ref{t4e3}, Equation \ref{t4e1} holds.
\end{proof}
\end{reptheorem}

We prove Theorems \ref{thm6} and \ref{thm7} first.
\begin{reptheorem}{thm6}
Suppose variable $X$ has $m$ values $x_1,...,x_m$ and $Y$ has $n$ values $y_1,...,y_n$, then the probability of causation $P({y_i}_{x_j}, x_k)$, where $1 \le i \le n, 1 \le j,k \le m, j\ne k$, is bounded as following:
\begin{eqnarray}
\max \left \{
\begin{array}{cc}
0, \\
P({y_i}_{x_j}) - P(x_j, y_i)\\
- 1 + P(x_j) + P(x_k) \\
\end{array}
\right \}
\le P({y_i}_{x_j}, x_k)
\label{t6e1}
\end{eqnarray}
\begin{eqnarray}
P({y_i}_{x_j}, x_k) \le \min \left \{
\begin{array}{cc}
 P({y_i}_{x_j}) - P(x_j, y_i), \\
 P(x_k) \\
\end{array} 
\right \}
\label{t6e2}
\end{eqnarray}
\begin{proof}
\begin{eqnarray}
&&P({y_i}_{x_j}, x_k)\nonumber\\
&=& P({y_i}_{x_j})-\sum_{t=1,t\ne k}^m P({y_i}_{x_j}, x_t)\nonumber\\
&=& P({y_i}_{x_j})-\sum_{t=1,t\ne k,j}^m P({y_i}_{x_j}, x_t)-P({y_i}_{x_j}, x_j)\nonumber\\
&=& P({y_i}_{x_j})-\sum_{t=1,t\ne k,j}^m P({y_i}_{x_j}, x_t)-P(x_j,y_i)\label{t6e3}\\
&\ge& P({y_i}_{x_j})-\sum_{t=1,t\ne k,j}^m P(x_t)-P(x_j,y_i)\nonumber\\
&=& P({y_i}_{x_j})-(1-P(x_k)-P(x_j))-P(x_j,y_i)\nonumber\\
&=& P({y_i}_{x_j})-P(x_j,y_i)-1+P(x_j)+P(x_k).\nonumber
\end{eqnarray}
Combine with $P({y_i}_{x_j}, x_k)\ge 0$, Equation \ref{t6e1} holds.\\
From Equation \ref{t6e3}, we also have,
\begin{eqnarray*}
&&P({y_i}_{x_j})-\sum_{t=1,t\ne k,j}^m P({y_i}_{x_j}, x_t)-P(x_j,y_i)\\
&\le& P({y_i}_{x_j})-P(x_j,y_i).
\end{eqnarray*}
Combine with $P({y_i}_{x_j}, x_k)\le P(x_k)$, Equation \ref{t6e2} holds.\\
\end{proof}
\end{reptheorem}

\begin{reptheorem}{thm7}
Suppose variable $X$ has $m$ values $x_1,...,x_m$ and $Y$ has $n$ values $y_1,...,y_n$, then the probability of causation $P({y_i}_{x_j}, y_k, x_p)$, where $1 \le i,k \le n, 1 \le j,p \le m, j\ne p$, is bounded as following:
\begin{eqnarray}
\max \left \{
\begin{array}{cc}
0, \\
 P({y_i}_{x_j}) + P(x_p, y_k)\\
- 1 + P(x_j) - P(x_j, y_i) \\
\end{array}
\right \}
\le P({y_i}_{x_j}, y_k, x_p)
\label{t7e1}
\end{eqnarray}
\begin{eqnarray}
P({y_i}_{x_j}, y_k, x_p) \le \min \left \{
\begin{array}{cc}
 P({y_i}_{x_j}) - P(x_j, y_i), \\
 P(x_p, y_k) \\
\end{array} 
\right \}
\label{t7e2}
\end{eqnarray}
\begin{proof}
\begin{eqnarray*}
&&P({y_i}_{x_j}, y_k, x_p)\\
&=& P({y_i}_{x_j})-\sum_{(t,r)\ne{p,k}} P({y_i}_{x_j}, x_t,y_r)\\
&=& P({y_i}_{x_j})-\sum_{(t,r)\ne(p,k),t\ne j} P({y_i}_{x_j}, x_t,y_r)-P(x_j,y_i)\\
&\ge& P({y_i}_{x_j})-\sum_{(t,r)\ne(p,k),t\ne j} P(x_t,y_r)-P(x_j,y_i)\\
&=& P({y_i}_{x_j})-(1-P(x_j)-P(x_p,y_k))-P(x_j,y_i)\\
&=& P({y_i}_{x_j})+P(x_p,y_k)-1+P(x_j)-P(x_j,y_i).
\end{eqnarray*}
Combine with $P({y_i}_{x_j}, y_k, x_p)\ge 0$, Equation \ref{t7e1} holds.\\
By Theorem \ref{thm6}, we also have,
\begin{eqnarray*}
&&P({y_i}_{x_j}, y_k, x_p)\\
&\le& P({y_i}_{x_j}, x_p)\\
&\le& P({y_i}_{x_j})-P(x_j,y_i).
\end{eqnarray*}
Combine with $P({y_i}_{x_j}, y_k, x_p)\le P(x_p, y_k)$, Equation \ref{t7e2} holds.\\
\end{proof}
\end{reptheorem}

Now we prove Theorem \ref{thm5}.
\begin{reptheorem}{thm5}
Suppose variable $X$ has $m$ values $x_1,...,x_m$ and $Y$ has $n$ values $y_1,...,y_n$, then the probability of causation $P({y_i}_{x_j}, y_k)$, where $1 \le i,k \le n, 1 \le j \le m, i\ne k$, is bounded as following:
\begin{eqnarray}
\max \left \{
\begin{array}{cc}
0, \\
P({y_i}_{x_j}) + P(y_k) - 1, \\
\sum_{1\le p\le m,p\ne j}\max \left \{
\begin{array}{cc}
0, \\
P({y_i}_{x_j})\\
+ P(x_p,y_k) \\
- 1 + P(x_j)\\
- P(x_j,y_i) \\
\end{array}
\right \}
\end{array}
\right \}\nonumber\\
\le P({y_i}_{x_j}, y_k)
\label{t5e1}
\end{eqnarray}
\begin{eqnarray}
P({y_i}_{x_j}, y_k) \le \min \left \{
\begin{array}{cc}
 P({y_i}_{x_j}) - P(x_j, y_i), \\
 P(y_k) - P(y_k, x_j) \\
\end{array} 
\right \}
\label{t5e2}
\end{eqnarray}
\begin{proof}
By Fréchet Inequalities, we have,
\begin{eqnarray*}
 &&P(A, B) \ge \max\{0, P(A)+P(B)-1\}.
\end{eqnarray*}
Thus, we have,
\begin{eqnarray}
 &&P({y_i}_{x_j}, y_k) \ge \max\{0, P({y_i}_{x_j}) + P(y_k) - 1\}.\label{t5e3}
\end{eqnarray}
By Theorem \ref{thm7}, we also have,
\begin{eqnarray*}
&&P({y_i}_{x_j}, y_k) \\
&=& \sum_{p=1,p\ne j}^m P({y_i}_{x_j}, y_k,x_p)\\
&\ge&\sum_{p=1,p\ne j}^m \max\{0,P({y_i}_{x_j})+P(x_p,y_k)\\
&&-1+P(x_j)-P(x_j,y_i)\}.
\end{eqnarray*}
Combine with Equation \ref{t5e3}, Equation \ref{t5e1} holds.\\
We also have,
\begin{eqnarray}
&&P({y_i}_{x_j}, y_k)\nonumber\\
&=& P({y_i}_{x_j}) - \sum_{r=1,r\ne k}^n P({y_i}_{x_j}, y_r)\nonumber\\
&\le& P({y_i}_{x_j}) - P({y_i}_{x_j}, y_i)\nonumber\\
&\le& P({y_i}_{x_j}) - P({y_i}_{x_j}, y_i,x_j)\nonumber\\
&=& P({y_i}_{x_j}) - P(x_j, y_i),\label{t5e4}
\end{eqnarray}
and,
\begin{eqnarray*}
&&P({y_i}_{x_j}, y_k) \\
&=& \sum_{p=1,p\ne j}^m P({y_i}_{x_j}, y_k,x_p)\\
&\le&\sum_{p=1,p\ne j}^m P(x_p,y_k)\\
&=&P(y_k)-P(y_k,x_j).
\end{eqnarray*}
Combine with Equation \ref{t5e4}, Equation \ref{t5e2} holds.
\end{proof}
\end{reptheorem}

\begin{reptheorem}{thm8}
Suppose variable $X$ has $m$ values $x_1,...,x_m$ and $Y$ has $n$ values $y_1,...,y_n$, then the probability of causation $P({y_{i_1}}_{x_{j_1}},...,{y_{i_k}}_{x_{j_k}})$, where $1 \le i_1,...,i_k \le n, 1 \le j_1,...,j_k \le m, j_1\ne ... \ne j_k$, is bounded as following:
\begin{eqnarray}
\max \left \{
\begin{array}{cc}
0, \\
\\
\sum_{1\le t\le k}P({y_{i_t}}_{x_{j_t}}) - k + 1, \\
\\
\max_{1\le t \le k} (LB(P({y_{i_1}}_{x_{j_1}},...,{y_{i_{t-1}}}_{x_{j_{t-1}}},\\
{y_{i_{t+1}}}_{x_{j_{t+1}}},...,{y_{i_k}}_{x_{j_k}}))\\
+ P({y_{i_t}}_{x_{i_t}}) - 1),\\
\\
\sum_{1\le p\le m, s.t., \exists r, 1\le r\le k, p=j_r }\\ LB(P({y_{i_1}}_{x_{j_1}},...,{y_{i_{r-1}}}_{x_{j_{r-1}}},\\
{y_{i_{r+1}}}_{x_{j_{r+1}}},...,{y_{i_k}}_{x_{j_k}}, x_{j_r}, y_{i_r})) + \\
\sum_{1\le p\le m, s.t., p \ne j_1 \ne ...\ne j_k}\\ LB(P({y_{i_1}}_{x_{j_1}},...,{y_{i_k}}_{x_{j_k}}, x_p))\\
\end{array}
\right \}\nonumber\\
\le P({y_{i_1}}_{x_{j_1}},...,{y_{i_k}}_{x_{j_k}})
\label{t8e1}
\end{eqnarray}
\begin{eqnarray}
P({y_{i_1}}_{x_{j_1}},...,{y_{i_k}}_{x_{j_k}}) \le \nonumber\\
\min \left \{
\begin{array}{cc}
 \min_{1\le t\le k} P({y_{i_t}}_{x_{j_t}}), \\
 \\
 \min_{1\le t \le k} UB(P({y_{i_1}}_{x_{j_1}},...,{y_{i_{t-1}}}_{x_{j_{t-1}}},\\
{y_{i_{t+1}}}_{x_{j_{t+1}}},...,{y_{i_k}}_{x_{j_k}})),\\
 \\
\sum_{1\le p\le m, s.t., \exists r, 1\le r\le k, p=j_r}\\ UB(P({y_{i_1}}_{x_{j_1}},...,{y_{i_{r-1}}}_{x_{j_{r-1}}},\\
{y_{i_{r+1}}}_{x_{j_{r+1}}},...,{y_{i_k}}_{x_{j_k}}, x_{j_r}, y_{i_r})) + \\
\sum_{1\le p\le m, s.t., p \ne j_1 \ne ...\ne j_k}\\ UB(P({y_{i_1}}_{x_{j_1}},...,{y_{i_k}}_{x_{j_k}}, x_p))\\
\end{array} 
\right \}
\label{t8e2}
\end{eqnarray}
where,\\
LB$(f)$ denotes the lower bound of a function $f$ and UB$(f)$ denotes the upper bound of a function $f$. The bounds of $P({y_{i_1}}_{x_{j_1}},...,{y_{i_{r-1}}}_{x_{j_{r-1}}},{y_{i_{r+1}}}_{x_{j_{r+1}}},...,{y_{i_k}}_{x_{j_k}}, x_{j_r}, y_{j_r})$ are given by Theorem \ref{thm7} or \ref{thm11}, the bounds of $P({y_{i_1}}_{x_{j_1}},...,{y_{i_k}}_{x_{j_k}}, x_p)$ are given by Theorem \ref{thm6} or \ref{thm9}, and the bounds of $P({y_{i_1}}_{x_{j_1}},...,{y_{i_{t-1}}}_{x_{j_{t-1}}},{y_{i_{t+1}}}_{x_{j_{t+1}}},...,{y_{i_k}}_{x_{j_k}})$ are given by Theorem \ref{thm8} or experimental data if $k=2$.
\begin{proof}
By Fréchet Inequalities, we have,
\begin{eqnarray*}
 P(A_1,...,A_n) &\ge& \max\{0, \\
 &&P(A_1)+...+P(A_n)-n+1\},\\
 P(A_1,...,A_n)&\le& \min\{P(A_1),...,P(A_n)\}.
\end{eqnarray*}
Thus, we have, 
\begin{eqnarray}
&&P({y_{i_1}}_{x_{j_1}},...,{y_{i_k}}_{x_{j_k}})\nonumber\\
&\ge& \max\{0,\nonumber\\
&&P({y_{i_1}}_{x_{j_1}})+...+P({y_{i_k}}_{x_{j_k}})-k+1\}\nonumber\\
&=&\max\{0, \sum_{1\le t\le k}P({y_{i_t}}_{x_{j_t}}) - k + 1\},\label{t8e3}\\
&&P({y_{i_1}}_{x_{j_1}},...,{y_{i_k}}_{x_{j_k}})\nonumber\\
&\le& \min\{P({y_{i_1}}_{x_{j_1}}),...,P({y_{i_k}}_{x_{j_k}})\}\nonumber\\
&=&\min_{1\le t\le k} P({y_{i_t}}_{x_{j_t}}).\label{t8e4}
\end{eqnarray}
Also by Fréchet Inequalities, we have,
\begin{eqnarray*}
 P(A,B)\ge P(A)+P(B)-1.
\end{eqnarray*}
Thus, we have,
\begin{eqnarray}
&&P({y_{i_1}}_{x_{j_1}},...,{y_{i_k}}_{x_{j_k}})\nonumber\\
&\ge& \max_{1\le t \le k}(P({y_{i_1}}_{x_{j_1}},...,{y_{i_{t-1}}}_{x_{j_{t-1}}},\nonumber\\
&&{y_{i_{t+1}}}_{x_{j_{t+1}}},...,{y_{i_k}}_{x_{j_k}})\nonumber\\
&&+ P({y_{i_t}}_{x_{i_t}}) - 1)\nonumber\\
&\ge& \max_{1\le t \le k}(LB(P({y_{i_1}}_{x_{j_1}},...,{y_{i_{t-1}}}_{x_{j_{t-1}}},\nonumber\\
&&{y_{i_{t+1}}}_{x_{j_{t+1}}},...,{y_{i_k}}_{x_{j_k}}))\nonumber\\
&&+ P({y_{i_t}}_{x_{i_t}}) - 1).\label{t8e5}
\end{eqnarray}
And, we have,
\begin{eqnarray}
&&P({y_{i_1}}_{x_{j_1}},...,{y_{i_k}}_{x_{j_k}})\nonumber\\
&\le& \min_{1\le t \le k}P({y_{i_1}}_{x_{j_1}},...,{y_{i_{t-1}}}_{x_{j_{t-1}}},\nonumber\\
&&{y_{i_{t+1}}}_{x_{j_{t+1}}},...,{y_{i_k}}_{x_{j_k}})\nonumber\\
&\le& \min_{1\le t \le k}UB(P({y_{i_1}}_{x_{j_1}},...,{y_{i_{t-1}}}_{x_{j_{t-1}}},\nonumber\\
&&{y_{i_{t+1}}}_{x_{j_{t+1}}},...,{y_{i_k}}_{x_{j_k}})).\label{t8e6}
\end{eqnarray}
Next, we have,
\begin{eqnarray}
&&P({y_{i_1}}_{x_{j_1}},...,{y_{i_k}}_{x_{j_k}})\nonumber\\
&=&\sum_{p=1}^m P({y_{i_1}}_{x_{j_1}},...,{y_{i_k}}_{x_{j_k}},x_p)\nonumber\\
&=&\sum_{1\le p\le m, s.t., \exists r, 1\le r\le k, p=j_r }\nonumber P({y_{i_1}}_{x_{j_1}},...,\nonumber\\
&&{y_{i_{r-1}}}_{x_{j_{r-1}}},{y_{i_{r+1}}}_{x_{j_{r+1}}},...,{y_{i_k}}_{x_{j_k}}, x_{j_r}, y_{i_r}) + \nonumber\\
&&\sum_{1\le p\le m, s.t., p \ne j_1 \ne ...\ne j_k}P({y_{i_1}}_{x_{j_1}},...,\nonumber\\
&&{y_{i_k}}_{x_{j_k}},x_p)\nonumber\\
&\ge&\sum_{1\le p\le m, s.t., \exists r, 1\le r\le k, p=j_r }\nonumber LB(P({y_{i_1}}_{x_{j_1}},...,\nonumber\\
&&{y_{i_{r-1}}}_{x_{j_{r-1}}},{y_{i_{r+1}}}_{x_{j_{r+1}}},...,{y_{i_k}}_{x_{j_k}}, x_{j_r}, y_{i_r})) + \nonumber\\
&&\sum_{1\le p\le m, s.t., p \ne j_1 \ne ...\ne j_k}LB(P({y_{i_1}}_{x_{j_1}},...,\nonumber\\
&&{y_{i_k}}_{x_{j_k}},x_p)).\label{t8e7}
\end{eqnarray}
And,
\begin{eqnarray}
&&P({y_{i_1}}_{x_{j_1}},...,{y_{i_k}}_{x_{j_k}})\nonumber\\
&=&\sum_{p=1}^m P({y_{i_1}}_{x_{j_1}},...,{y_{i_k}}_{x_{j_k}},x_p)\nonumber\\
&=&\sum_{1\le p\le m, s.t., \exists r, 1\le r\le k, p=j_r }\nonumber P({y_{i_1}}_{x_{j_1}},...,\nonumber\\
&&{y_{i_{r-1}}}_{x_{j_{r-1}}},{y_{i_{r+1}}}_{x_{j_{r+1}}},...,{y_{i_k}}_{x_{j_k}}, x_{j_r}, y_{i_r}) + \nonumber\\
&&\sum_{1\le p\le m, s.t., p \ne j_1 \ne ...\ne j_k}P({y_{i_1}}_{x_{j_1}},...,\nonumber\\
&&{y_{i_k}}_{x_{j_k}},x_p)\nonumber\\
&\le&\sum_{1\le p\le m, s.t., \exists r, 1\le r\le k, p=j_r }\nonumber UB(P({y_{i_1}}_{x_{j_1}},...,\nonumber\\
&&{y_{i_{r-1}}}_{x_{j_{r-1}}},{y_{i_{r+1}}}_{x_{j_{r+1}}},...,{y_{i_k}}_{x_{j_k}}, x_{j_r}, y_{i_r})) + \nonumber\\
&&\sum_{1\le p\le m, s.t., p \ne j_1 \ne ...\ne j_k}UB(P({y_{i_1}}_{x_{j_1}},...,\nonumber\\
&&{y_{i_k}}_{x_{j_k}},x_p)).\label{t8e8}
\end{eqnarray}
Combine Equations \ref{t8e3}, \ref{t8e5}, and \ref{t8e7}, Equation \ref{t8e1} holds and Combine Equations \ref{t8e4}, \ref{t8e6}, and \ref{t8e8}, Equation \ref{t8e2} holds. 
\end{proof}
\end{reptheorem}

\begin{reptheorem}{thm9}
Suppose variable $X$ has $m$ values $x_1,...,x_m$ and $Y$ has $n$ values $y_1,...,y_n$, then the probability of causation $P({y_{i_1}}_{x_{j_1}},...,{y_{i_k}}_{x_{j_k}},x_p)$, where $1 \le i_1,...,i_k \le n, 1 \le j_1,...,j_k,p \le m, j_1\ne ... \ne j_k \ne p$, is bounded as following:
\begin{eqnarray}
\max \left \{
\begin{array}{cc}
0, \\
\\
\sum_{1\le t\le k}P({y_{i_t}}_{x_{j_t}}) +P(x_p) - k, \\
\\
\max_{1\le t \le k} (LB(P({y_{i_1}}_{x_{j_1}},...,{y_{i_{t-1}}}_{x_{j_{t-1}}},\\
{y_{i_{t+1}}}_{x_{j_{t+1}}},...,{y_{i_k}}_{x_{j_k}}))\\
+ LB(P({y_{i_t}}_{x_{i_t}},x_p)) - 1)\\
\end{array}
\right \}\nonumber\\
\le P({y_{i_1}}_{x_{j_1}},...,{y_{i_k}}_{x_{j_k}},x_p)
\label{t9e1}
\end{eqnarray}
\begin{eqnarray}
P({y_{i_1}}_{x_{j_1}},...,{y_{i_k}}_{x_{j_k}},x_p) \le \nonumber\\
\min \left \{
\begin{array}{cc}
 \min_{1\le t\le k} P({y_{i_t}}_{x_{j_t}}), \\
 \\
 P(x_p),\\
 \\
\min_{1\le t \le k} UB(P({y_{i_1}}_{x_{j_1}},...,{y_{i_{t-1}}}_{x_{j_{t-1}}},\\
{y_{i_{t+1}}}_{x_{j_{t+1}}},...,{y_{i_k}}_{x_{j_k}})),\\
\\
\min_{1\le t \le k} UB(P({y_{i_t}}_{x_{i_t}},x_p))\\
\end{array} 
\right \}
\label{t9e2}
\end{eqnarray}
where,\\
LB$(f)$ denotes the lower bound of a function $f$ and UB$(f)$ denotes the upper bound of a function $f$. The bounds of $P({y_{i_1}}_{x_{j_1}},...,{y_{i_{t-1}}}_{x_{j_{t-1}}},{y_{i_{t+1}}}_{x_{j_{t+1}}},...,{y_{i_k}}_{x_{j_k}})$ are given by Theorem \ref{thm8} or experimental data if $k=2$ and the bounds of $P({y_{i_t}}_{x_{i_t}},x_p)$ are given by Theorem \ref{thm6}.
\begin{proof}
By Fréchet Inequalities, we have,
\begin{eqnarray*}
 P(A_1,...,A_n) &\ge& \max\{0, \\
 &&P(A_1)+...+P(A_n)-n+1\},\\
 P(A_1,...,A_n)&\le& \min\{P(A_1),...,P(A_n)\}.
\end{eqnarray*}
Thus, we have, 
\begin{eqnarray}
&&P({y_{i_1}}_{x_{j_1}},...,{y_{i_k}}_{x_{j_k}},x_p)\nonumber\\
&\ge& \max\{0,\nonumber\\
&&P({y_{i_1}}_{x_{j_1}})+...+P({y_{i_k}}_{x_{j_k}})+P(x_p)-k\}\nonumber\\
&=&\max\{0, \sum_{1\le t\le k}P({y_{i_t}}_{x_{j_t}}) +P(x_p) - k\},\label{t9e3}\\
&&P({y_{i_1}}_{x_{j_1}},...,{y_{i_k}}_{x_{j_k}},x_p)\nonumber\\
&\le& \min\{P({y_{i_1}}_{x_{j_1}}),...,P({y_{i_k}}_{x_{j_k}}),P(x_p)\}\nonumber\\
&=&\min\{\min_{1\le t\le k} P({y_{i_t}}_{x_{j_t}}),P(x_p)\}.\label{t9e4}
\end{eqnarray}
Also by Fréchet Inequalities, we have,
\begin{eqnarray*}
 P(A,B)\ge P(A)+P(B)-1,\\
 P(A,B)\le \min\{P(A),P(B)\}.
\end{eqnarray*}
Thus, we have,
\begin{eqnarray}
&&P({y_{i_1}}_{x_{j_1}},...,{y_{i_k}}_{x_{j_k}},x_p)\nonumber\\
&\ge& \max_{1\le t \le k}(P({y_{i_1}}_{x_{j_1}},...,{y_{i_{t-1}}}_{x_{j_{t-1}}},\nonumber\\
&&{y_{i_{t+1}}}_{x_{j_{t+1}}},...,{y_{i_k}}_{x_{j_k}})\nonumber\\
&&+ P({y_{i_t}}_{x_{i_t}},x_p) - 1)\nonumber\\
&\ge& \max_{1\le t \le k}(LB(P({y_{i_1}}_{x_{j_1}},...,{y_{i_{t-1}}}_{x_{j_{t-1}}},\nonumber\\
&&{y_{i_{t+1}}}_{x_{j_{t+1}}},...,{y_{i_k}}_{x_{j_k}}))\nonumber\\
&&+ LB(P({y_{i_t}}_{x_{i_t}},x_p)) - 1).\label{t9e5}
\end{eqnarray}
And, we have,
\begin{eqnarray}
&&P({y_{i_1}}_{x_{j_1}},...,{y_{i_k}}_{x_{j_k}},x_p)\nonumber\\
&\le& \min_{1\le t \le k}\min\{P({y_{i_1}}_{x_{j_1}},...,{y_{i_{t-1}}}_{x_{j_{t-1}}},\nonumber\\
&&{y_{i_{t+1}}}_{x_{j_{t+1}}},...,{y_{i_k}}_{x_{j_k}}),P({y_{i_t}}_{x_{j_t}},x_p)\}\nonumber\\
&\le& \min_{1\le t \le k}\min\{UB(P({y_{i_1}}_{x_{j_1}},...,{y_{i_{t-1}}}_{x_{j_{t-1}}},\nonumber\\
&&{y_{i_{t+1}}}_{x_{j_{t+1}}},...,{y_{i_k}}_{x_{j_k}})),UB(P({y_{i_t}}_{x_{j_t}},x_p))\}.\label{t9e6}
\end{eqnarray}
Combine Equations \ref{t9e3} and \ref{t9e5}, Equation \ref{t9e1} holds and Combine Equations \ref{t9e4} and \ref{t9e6}, Equation \ref{t9e2} holds. 
\end{proof}
\end{reptheorem}

\begin{reptheorem}{thm10}
Suppose variable $X$ has $m$ values $x_1,...,x_m$ and $Y$ has $n$ values $y_1,...,y_n$, then the probability of causation $P({y_{i_1}}_{x_{j_1}},...,{y_{i_k}}_{x_{j_k}},y_q)$, where $1 \le i_1,...,i_k,q \le n, 1 \le j_1,...,j_k \le m, j_1\ne ... \ne j_k$, is bounded as following:
\begin{eqnarray}
\max \left \{
\begin{array}{cc}
0,\\
\\
\sum_{1\le t\le k}P({y_{i_t}}_{x_{j_t}}) +P(y_q) - k, \\
\\
\max_{1\le t \le k} (LB(P({y_{i_1}}_{x_{j_1}},...,{y_{i_{t-1}}}_{x_{j_{t-1}}},\\
{y_{i_{t+1}}}_{x_{j_{t+1}}},...,{y_{i_k}}_{x_{j_k}}))\\
+ LB(P({y_{i_t}}_{x_{i_t}},y_q)) - 1),\\
\\
\sum_{1\le p\le m, \exists r, 1\le r\le k, p=j_r, q=i_r}\\ LB(P({y_{i_1}}_{x_{j_1}},...,{y_{i_{r-1}}}_{x_{j_{r-1}}},\\
{y_{i_{r+1}}}_{x_{j_{r+1}}},...,{y_{i_k}}_{x_{j_k}}, x_{j_r}, y_{i_r})) + \\
\sum_{1\le p\le m, s.t., p \ne j_1 \ne ...\ne j_k}\\ LB(P({y_{i_1}}_{x_{j_1}},...,{y_{i_k}}_{x_{j_k}}, x_p, y_q))\\
\end{array}
\right \}\nonumber\\
\le P({y_{i_1}}_{x_{j_1}},...,{y_{i_k}}_{x_{j_k}}, y_q)
\label{t10e1}
\end{eqnarray}
\begin{eqnarray}
P({y_{i_1}}_{x_{j_1}},...,{y_{i_k}}_{x_{j_k}},y_q) \le \nonumber\\
\min \left \{
\begin{array}{cc}
 \min_{1\le t\le k} P({y_{i_t}}_{x_{j_t}}), \\
 \\
 P(y_q),\\
 \\
\min_{1\le t \le k} UB(P({y_{i_1}}_{x_{j_1}},...,{y_{i_{t-1}}}_{x_{j_{t-1}}},\\
{y_{i_{t+1}}}_{x_{j_{t+1}}},...,{y_{i_k}}_{x_{j_k}})),\\
\\
\min_{1\le t \le k} UB(P({y_{i_t}}_{x_{i_t}},y_q)),\\
\\
\sum_{1\le p\le m, s.t., \exists r, 1\le r\le k, p=j_r, q=i_r}\\ UB(P({y_{i_1}}_{x_{j_1}},...,{y_{i_{r-1}}}_{x_{j_{r-1}}},\\
{y_{i_{r+1}}}_{x_{j_{r+1}}},...,{y_{i_k}}_{x_{j_k}}, x_{j_r}, y_{i_r})) + \\
\sum_{1\le p\le m, s.t., p \ne j_1 \ne ...\ne j_k}\\ UB(P({y_{i_1}}_{x_{j_1}},...,{y_{i_k}}_{x_{j_k}}, x_p, y_q))\\
\end{array} 
\right \}
\label{t10e2}
\end{eqnarray}
where,\\
LB$(f)$ denotes the lower bound of a function $f$ and UB$(f)$ denotes the upper bound of a function $f$. The bounds of $P({y_{i_1}}_{x_{j_1}},...,{y_{i_{r-1}}}_{x_{j_{r-1}}},{y_{i_{r+1}}}_{x_{j_{r+1}}},...,{y_{i_k}}_{x_{j_k}}, x_{j_r}, y_{j_r})$, $P({y_{i_1}}_{x_{j_1}},...,{y_{i_k}}_{x_{j_k}}, x_p, y_q)$ are given by Theorem \ref{thm7} or \ref{thm11}, the bounds of $P({y_{i_1}}_{x_{j_1}},...,{y_{i_{t-1}}}_{x_{j_{t-1}}}, {y_{i_{t+1}}}_{x_{j_{t+1}}},...,{y_{i_k}}_{x_{j_k}})$ are given by Theorem \ref{thm8} or experimental data if $k=2$, and the bounds of $P({y_{i_t}}_{x_{i_t}},y_q)$ are given by Theorem \ref{thm4} or \ref{thm5}.
\begin{proof}
By Fréchet Inequalities, we have,
\begin{eqnarray*}
 P(A_1,...,A_n) &\ge& \max\{0, \\
 &&P(A_1)+...+P(A_n)-n+1\},\\
 P(A_1,...,A_n)&\le& \min\{P(A_1),...,P(A_n)\}.
\end{eqnarray*}
Thus, we have, 
\begin{eqnarray}
&&P({y_{i_1}}_{x_{j_1}},...,{y_{i_k}}_{x_{j_k}},y_q)\nonumber\\
&\ge& \max\{0,\nonumber\\
&&P({y_{i_1}}_{x_{j_1}})+...+P({y_{i_k}}_{x_{j_k}})+P(y_q)-k\}\nonumber\\
&=&\max\{0, \sum_{1\le t\le k}P({y_{i_t}}_{x_{j_t}}) +P(y_q)- k\},\label{t10e3}\\
&&P({y_{i_1}}_{x_{j_1}},...,{y_{i_k}}_{x_{j_k}},y_q)\nonumber\\
&\le& \min\{P({y_{i_1}}_{x_{j_1}}),...,P({y_{i_k}}_{x_{j_k}}),P(y_q)\}\nonumber\\
&=&\min\{\min_{1\le t\le k} P({y_{i_t}}_{x_{j_t}}),P(y_q)\}.\label{t10e4}
\end{eqnarray}
Also by Fréchet Inequalities, we have,
\begin{eqnarray*}
 P(A,B)\ge P(A)+P(B)-1,\\
 P(A,B)\le \min\{P(A),P(B)\}.
\end{eqnarray*}
Thus, we have,
\begin{eqnarray}
&&P({y_{i_1}}_{x_{j_1}},...,{y_{i_k}}_{x_{j_k}},y_q)\nonumber\\
&\ge& \max_{1\le t \le k}(P({y_{i_1}}_{x_{j_1}},...,{y_{i_{t-1}}}_{x_{j_{t-1}}},\nonumber\\
&&{y_{i_{t+1}}}_{x_{j_{t+1}}},...,{y_{i_k}}_{x_{j_k}})\nonumber\\
&&+ P({y_{i_t}}_{x_{i_t}},y_q) - 1)\nonumber\\
&\ge& \max_{1\le t \le k}(LB(P({y_{i_1}}_{x_{j_1}},...,{y_{i_{t-1}}}_{x_{j_{t-1}}},\nonumber\\
&&{y_{i_{t+1}}}_{x_{j_{t+1}}},...,{y_{i_k}}_{x_{j_k}}))\nonumber\\
&&+ LB(P({y_{i_t}}_{x_{i_t}},y_q)) - 1).\label{t10e5}
\end{eqnarray}
And, we have,
\begin{eqnarray}
&&P({y_{i_1}}_{x_{j_1}},...,{y_{i_k}}_{x_{j_k}},y_q)\nonumber\\
&\le& \min_{1\le t \le k}\min\{P({y_{i_1}}_{x_{j_1}},...,{y_{i_{t-1}}}_{x_{j_{t-1}}},\nonumber\\
&&{y_{i_{t+1}}}_{x_{j_{t+1}}},...,{y_{i_k}}_{x_{j_k}}),P({y_{i_t}}_{x_{j_t}},y_q)\}\nonumber\\
&\le& \min_{1\le t \le k}\min\{UB(P({y_{i_1}}_{x_{j_1}},...,{y_{i_{t-1}}}_{x_{j_{t-1}}},\nonumber\\
&&{y_{i_{t+1}}}_{x_{j_{t+1}}},...,{y_{i_k}}_{x_{j_k}})),UB(P({y_{i_t}}_{x_{j_t}},y_q))\}.\label{t10e6}
\end{eqnarray}
Next, we have,
\begin{eqnarray}
&&P({y_{i_1}}_{x_{j_1}},...,{y_{i_k}}_{x_{j_k}},y_q)\nonumber\\
&=&\sum_{p=1}^m P({y_{i_1}}_{x_{j_1}},...,{y_{i_k}}_{x_{j_k}},y_q,x_p)\nonumber\\
&=&\sum_{1\le p\le m, s.t., \exists r, 1\le r\le k, p=j_r, q=i_r }\nonumber P({y_{i_1}}_{x_{j_1}},...,\nonumber\\
&&{y_{i_{r-1}}}_{x_{j_{r-1}}},{y_{i_{r+1}}}_{x_{j_{r+1}}},...,{y_{i_k}}_{x_{j_k}}, x_{j_r}, y_{i_r}) + \nonumber\\
&&\sum_{1\le p\le m, s.t., p \ne j_1 \ne ...\ne j_k}P({y_{i_1}}_{x_{j_1}},...,\nonumber\\
&&{y_{i_k}}_{x_{j_k}},x_p,y_q)\nonumber\\
&\ge&\sum_{1\le p\le m, s.t., \exists r, 1\le r\le k, p=j_r, q=i_r}\nonumber LB(P({y_{i_1}}_{x_{j_1}},...,\nonumber\\
&&{y_{i_{r-1}}}_{x_{j_{r-1}}},{y_{i_{r+1}}}_{x_{j_{r+1}}},...,{y_{i_k}}_{x_{j_k}}, x_{j_r}, y_{i_r})) + \nonumber\\
&&\sum_{1\le p\le m, s.t., p \ne j_1 \ne ...\ne j_k}LB(P({y_{i_1}}_{x_{j_1}},...,\nonumber\\
&&{y_{i_k}}_{x_{j_k}},x_p,y_q)).\label{t10e7}
\end{eqnarray}
And,
\begin{eqnarray}
&&P({y_{i_1}}_{x_{j_1}},...,{y_{i_k}}_{x_{j_k}},y_q)\nonumber\\
&=&\sum_{p=1}^m P({y_{i_1}}_{x_{j_1}},...,{y_{i_k}}_{x_{j_k}},y_q,x_p)\nonumber\\
&=&\sum_{1\le p\le m, s.t., \exists r, 1\le r\le k, p=j_r, q=i_r}\nonumber P({y_{i_1}}_{x_{j_1}},...,\nonumber\\
&&{y_{i_{r-1}}}_{x_{j_{r-1}}},{y_{i_{r+1}}}_{x_{j_{r+1}}},...,{y_{i_k}}_{x_{j_k}}, x_{j_r}, y_{i_r}) + \nonumber\\
&&\sum_{1\le p\le m, s.t., p \ne j_1 \ne ...\ne j_k}P({y_{i_1}}_{x_{j_1}},...,\nonumber\\
&&{y_{i_k}}_{x_{j_k}},x_p,y_q)\nonumber\\
&\le&\sum_{1\le p\le m, s.t., \exists r, 1\le r\le k, p=j_r, q=i_r}\nonumber UB(P({y_{i_1}}_{x_{j_1}},...,\nonumber\\
&&{y_{i_{r-1}}}_{x_{j_{r-1}}},{y_{i_{r+1}}}_{x_{j_{r+1}}},...,{y_{i_k}}_{x_{j_k}}, x_{j_r}, y_{i_r})) + \nonumber\\
&&\sum_{1\le p\le m, s.t., p \ne j_1 \ne ...\ne j_k}UB(P({y_{i_1}}_{x_{j_1}},...,\nonumber\\
&&{y_{i_k}}_{x_{j_k}},x_p,y_q)).\label{t10e8}
\end{eqnarray}
Combine Equations \ref{t10e3}, \ref{t10e5}, and \ref{t10e7}, Equation \ref{t10e1} holds and Combine Equations \ref{t10e4}, \ref{t10e6}, and \ref{t10e8}, Equation \ref{t10e2} holds. 
\end{proof}
\end{reptheorem}

\begin{reptheorem}{thm11}
Suppose variable $X$ has $m$ values $x_1,...,x_m$ and $Y$ has $n$ values $y_1,...,y_n$, then the probability of causation $P({y_{i_1}}_{x_{j_1}},...,{y_{i_k}}_{x_{j_k}},x_p,y_q)$, where $1 \le i_1,...,i_k,q \le n, 1 \le j_1,...,j_k,p \le m, j_1\ne ... \ne j_k \ne p$, is bounded as following:
\begin{eqnarray}
\max \left \{
\begin{array}{cc}
0, \\
\\
\sum_{1\le t\le k}P({y_{i_t}}_{x_{j_t}}) +P(x_p,y_q) - k, \\
\\
\max_{1\le t \le k} (LB(P({y_{i_1}}_{x_{j_1}},...,{y_{i_{t-1}}}_{x_{j_{t-1}}},\\
{y_{i_{t+1}}}_{x_{j_{t+1}}},...,{y_{i_k}}_{x_{j_k}}))\\
+ LB(P({y_{i_t}}_{x_{i_t}},x_p,y_q)) - 1)\\
\end{array}
\right \}\nonumber\\
\le P({y_{i_1}}_{x_{j_1}},...,{y_{i_k}}_{x_{j_k}},x_p,y_q)
\label{t11e1}
\end{eqnarray}
\begin{eqnarray}
P({y_{i_1}}_{x_{j_1}},...,{y_{i_k}}_{x_{j_k}},x_p,y_q) \le \nonumber\\
\min \left \{
\begin{array}{cc}
 \min_{1\le t\le k} P({y_{i_t}}_{x_{j_t}}), \\
 \\
 P(x_p,y_q),\\
 \\
\min_{1\le t \le k} UB(P({y_{i_1}}_{x_{j_1}},...,{y_{i_{t-1}}}_{x_{j_{t-1}}},\\
{y_{i_{t+1}}}_{x_{j_{t+1}}},...,{y_{i_k}}_{x_{j_k}})),\\
\\
\min_{1\le t \le k} UB(P({y_{i_t}}_{x_{i_t}},x_p,y_q))\\
\end{array} 
\right \}
\label{t11e2}
\end{eqnarray}
where,\\
LB$(f)$ denotes the lower bound of a function $f$ and UB$(f)$ denotes the upper bound of a function $f$. The bounds of $P({y_{i_1}}_{x_{j_1}},...,{y_{i_{t-1}}}_{x_{j_{t-1}}}, {y_{i_{t+1}}}_{x_{j_{t+1}}},...,{y_{i_k}}_{x_{j_k}})$ are given by Theorem \ref{thm8} or experimental data if $k=2$ and the bounds of $P({y_{i_t}}_{x_{i_t}},x_p,y_q)$ are given by Theorem \ref{thm7}.
\begin{proof}
By Fréchet Inequalities, we have,
\begin{eqnarray*}
 P(A_1,...,A_n) &\ge& \max\{0, \\
 &&P(A_1)+...+P(A_n)-n+1\},\\
 P(A_1,...,A_n)&\le& \min\{P(A_1),...,P(A_n)\}.
\end{eqnarray*}
Thus, we have, 
\begin{eqnarray}
&&P({y_{i_1}}_{x_{j_1}},...,{y_{i_k}}_{x_{j_k}},x_p,y_q)\nonumber\\
&\ge& \max\{0,\nonumber\\
&&P({y_{i_1}}_{x_{j_1}})+...+P({y_{i_k}}_{x_{j_k}})+P(x_p,y_q)-k\}\nonumber\\
&=&\max\{0, \sum_{1\le t\le k}P({y_{i_t}}_{x_{j_t}}) +P(x_p,y_q) - k\},\label{t11e3}\\
&&P({y_{i_1}}_{x_{j_1}},...,{y_{i_k}}_{x_{j_k}},x_p,y_q)\nonumber\\
&\le& \min\{P({y_{i_1}}_{x_{j_1}}),...,P({y_{i_k}}_{x_{j_k}}),P(x_p)\}\nonumber\\
&=&\min\{\min_{1\le t\le k} P({y_{i_t}}_{x_{j_t}}),P(x_p,y_q)\}.\label{t11e4}
\end{eqnarray}
Also by Fréchet Inequalities, we have,
\begin{eqnarray*}
 P(A,B)\ge P(A)+P(B)-1,\\
 P(A,B)\le \min\{P(A),P(B)\}.
\end{eqnarray*}
Thus, we have,
\begin{eqnarray}
&&P({y_{i_1}}_{x_{j_1}},...,{y_{i_k}}_{x_{j_k}},x_p,y_q)\nonumber\\
&\ge& \max_{1\le t \le k}(P({y_{i_1}}_{x_{j_1}},...,{y_{i_{t-1}}}_{x_{j_{t-1}}},\nonumber\\
&&{y_{i_{t+1}}}_{x_{j_{t+1}}},...,{y_{i_k}}_{x_{j_k}})\nonumber\\
&&+ P({y_{i_t}}_{x_{i_t}},x_p,y_q) - 1)\nonumber\\
&\ge& \max_{1\le t \le k}(LB(P({y_{i_1}}_{x_{j_1}},...,{y_{i_{t-1}}}_{x_{j_{t-1}}},\nonumber\\
&&{y_{i_{t+1}}}_{x_{j_{t+1}}},...,{y_{i_k}}_{x_{j_k}}))\nonumber\\
&&+ LB(P({y_{i_t}}_{x_{i_t}},x_p,y_q)) - 1).\label{t11e5}
\end{eqnarray}
And, we have,
\begin{eqnarray}
&&P({y_{i_1}}_{x_{j_1}},...,{y_{i_k}}_{x_{j_k}},x_p,y_q)\nonumber\\
&\le& \min_{1\le t \le k}\min\{P({y_{i_1}}_{x_{j_1}},...,{y_{i_{t-1}}}_{x_{j_{t-1}}},\nonumber\\
&&{y_{i_{t+1}}}_{x_{j_{t+1}}},...,{y_{i_k}}_{x_{j_k}}),P({y_{i_t}}_{x_{j_t}},x_p,y_q)\}\nonumber\\
&\le& \min_{1\le t \le k}\min\{UB(P({y_{i_1}}_{x_{j_1}},...,{y_{i_{t-1}}}_{x_{j_{t-1}}},\nonumber\\
&&{y_{i_{t+1}}}_{x_{j_{t+1}}},...,{y_{i_k}}_{x_{j_k}})),\nonumber\\
&&UB(P({y_{i_t}}_{x_{j_t}},x_p,y_q))\}.\label{t11e6}
\end{eqnarray}
Combine Equations \ref{t11e3} and \ref{t11e5}, Equation \ref{t11e1} holds and Combine Equations \ref{t11e4} and \ref{t11e6}, Equation \ref{t11e2} holds. 
\end{proof}
\end{reptheorem}

\subsection{Calculation in the Examples}
\subsubsection{Choice of Treatment}
First, by Theorem \ref{thm8},
\begin{eqnarray*}
&&P({y_3}_{x_1},{y_1}_{x_2},{y_2}_{x_3}) \\
&\ge& \max\{P({y_3}_{x_1})+P({y_1}_{x_2})+P({y_2}_{x_3})-2,0\}\\
&=&\max\{213/300+184/300+189/300-2,0\}\\
&=&0.
\end{eqnarray*}
and,
\begin{eqnarray*}
&&P({y_3}_{x_1},{y_1}_{x_2},{y_2}_{x_3}) \\
&\le& \min\{P({y_3}_{x_1}),P({y_1}_{x_2}),P({y_2}_{x_3})\}\\
&=&\min\{213/300,184/300,189/300\}\\
&=&0.613.
\end{eqnarray*}
Second, by Theorem \ref{thm8}, we have,
\begin{eqnarray*}
0.323 \le P({y_3}_{x_1},{y_1}_{x_2})\le 0.340,\\
0.243 \le P({y_1}_{x_2},{y_2}_{x_3})\le 0.386,\\
0.340 \le P({y_3}_{x_1},{y_2}_{x_3})\le 0.472.\\
\end{eqnarray*}
Then, by Theorem \ref{thm8} again,
\begin{eqnarray*}
&&P({y_3}_{x_1},{y_1}_{x_2},{y_2}_{x_3}) \\
&\ge& \max\{LB(P({y_3}_{x_1},{y_1}_{x_2}))+P({y_2}_{x_3})-1,\\
&&LB(P({y_1}_{x_2},{y_2}_{x_3}))+P({y_3}_{x_1})-1,\\
&&LB(P({y_3}_{x_1},{y_2}_{x_3}))+P({y_1}_{x_2})-1\}\\
&=&\max\{0.323+189/300-1,\\
&&0.243+213/300-1,\\
&&0.340+184/300-1\}\\
&=&-0.047.
\end{eqnarray*}
and,
\begin{eqnarray*}
&&P({y_3}_{x_1},{y_1}_{x_2},{y_2}_{x_3}) \\
&\le& \min\{LB(P({y_3}_{x_1},{y_1}_{x_2})),\\
&&LB(P({y_1}_{x_2},{y_2}_{x_3})),\\
&&LB(P({y_3}_{x_1},{y_2}_{x_3}))\}\\
&=&\min\{0.340,0.386,0.472\}\\
&=&0.340.
\end{eqnarray*}
Third, by Theorem \ref{thm11}, we have,
\begin{eqnarray*}
0 \le P(x_1,y_3,{y_1}_{x_2},{y_2}_{x_3})\le 0.008,\\
0 \le P(x_2,y_1,{y_3}_{x_1},{y_2}_{x_3})\le 0.011,\\
0 \le P(x_3,y_2,{y_3}_{x_1},{y_1}_{x_2})\le 0.080.\\
\end{eqnarray*}
Then, by Theorem \ref{thm8},
\begin{eqnarray*}
&&P({y_3}_{x_1},{y_1}_{x_2},{y_2}_{x_3}) \\
&\ge& LB(P(x_1,y_3,{y_1}_{x_2},{y_2}_{x_3}))+\\
&&LB(P(x_2,y_1,{y_3}_{x_1},{y_2}_{x_3}))+\\
&&LB(P(x_3,y_2,{y_3}_{x_1},{y_1}_{x_2}))\\
&=& 0.
\end{eqnarray*}
and,
\begin{eqnarray*}
&&P({y_3}_{x_1},{y_1}_{x_2},{y_2}_{x_3}) \\
&\le& UB(P(x_1,y_3,{y_1}_{x_2},{y_2}_{x_3}))+\\
&&UB(P(x_2,y_1,{y_3}_{x_1},{y_2}_{x_3}))+\\
&&UB(P(x_3,y_2,{y_3}_{x_1},{y_1}_{x_2}))\\
&=& 0.099.
\end{eqnarray*}
Combine all the bounds of $P({y_3}_{x_1},{y_1}_{x_2},{y_2}_{x_3})$, we then have,
\begin{eqnarray*}
&&\max\{0,-0.047,0\}\le P({y_3}_{x_1},{y_1}_{x_2},{y_2}_{x_3}),\\
&&P({y_3}_{x_1},{y_1}_{x_2},{y_2}_{x_3}) \le \min\{0.613,0.340,0.099\}
\end{eqnarray*}
Finally, we obtain,
\begin{eqnarray*}
0\le P({y_3}_{x_1},{y_1}_{x_2},{y_2}_{x_3}) \le 0.099.
\end{eqnarray*}

\subsubsection{Change of Institute}
First, by Theorem \ref{thm7},
\begin{eqnarray*}
&&P({y_1}_{x_3},x_2,y_2)\\
&\ge &\max\{0,\\
&&P({y_1}_{x_3})+P(x_2,y_2)-1+P(x_3)-P(x_3,y_1)\}\\
&=&\max\{0,\\
&&234/300+118/1200-1+255/1200-24/1200\}\\
&=&85/1200.
\end{eqnarray*}
and,
\begin{eqnarray*}
&&P({y_1}_{x_3},x_2,y_2)\\
&\le &\min\{P({y_1}_{x_3})-P(x_3,y_1),P(x_2,y_2)\}\\
&=&\min\{234/300-24/1200,118/1200\}\\
&=&118/1200.
\end{eqnarray*}
Also,
\begin{eqnarray*}
P({y_1}_{x_3}|x_2,y_2)=P({y_1}_{x_3},x_2,y_2)/P(x_2,y_2).
\end{eqnarray*}
Thus,
\begin{eqnarray*}
&&\frac{85/1200}{118/1200}\le P({y_1}_{x_3}|x_2,y_2)\le \frac{118/1200}{118/1200},\\
&&0.720 \le P({y_1}_{x_3}|x_2,y_2)\le 1.
\end{eqnarray*}
Again, by Theorem \ref{thm7},
\begin{eqnarray*}
&&P({y_1}_{x_4},x_2,y_2)\\
&\ge &\max\{0,\\
&&P({y_1}_{x_4})+P(x_2,y_2)-1+P(x_4)-P(x_4,y_1)\}\\
&=&\max\{0,\\
&&151/300+118/1200-1+622/1200-599/1200\}\\
&=&0.
\end{eqnarray*}
and,
\begin{eqnarray*}
&&P({y_1}_{x_4},x_2,y_2)\\
&\le &\min\{P({y_1}_{x_4})-P(x_4,y_1),P(x_2,y_2)\}\\
&=&\min\{151/300-599/1200,118/1200\}\\
&=&5/1200.
\end{eqnarray*}
Also,
\begin{eqnarray*}
P({y_1}_{x_4}|x_2,y_2)=P({y_1}_{x_4},x_2,y_2)/P(x_2,y_2).
\end{eqnarray*}
Thus,
\begin{eqnarray*}
&&\frac{0}{118/1200}\le P({y_1}_{x_4}|x_2,y_2)\le \frac{5/1200}{118/1200},\\
&&0 \le P({y_1}_{x_4}|x_2,y_2)\le 0.042.
\end{eqnarray*}

\subsubsection{Effectiveness of Vaccine}
First, by Theorem \ref{thm7}, we have the following,
\begin{eqnarray*}
0\le P({y_4}_{x_2},x_1,y_1)\le 0.005,\\
0\le P({y_1}_{x_1},x_2,y_4)\le 0.034,\\
0.037\le P({y_4}_{x_2},x_1,y_2)\le 0.062,\\
0\le P({y_2}_{x_1},x_2,y_4)\le 0.015,\\
0.502\le P({y_4}_{x_2},x_1,y_3)\le 0.527,\\
0\le P({y_3}_{x_1},x_2,y_4)\le 0.034.\\
\end{eqnarray*}
Then by Theorem \ref{thm8},
\begin{eqnarray*}
&&P({y_1}_{x_1},{y_4}_{x_2})\\
&\ge&\max\{0, P({y_1}_{x_1})+P({y_4}_{x_2})-1,\\
&&LB(P({y_4}_{x_2},x_1,y_1))+LB(P({y_1}_{x_1},x_2,y_4))\}\\
&=&\max\{0,205/600+364/600-1,0+0\}\\
&=&0.
\end{eqnarray*}
and,
\begin{eqnarray*}
&&P({y_1}_{x_1},{y_4}_{x_2})\\
&\le&\min\{P({y_1}_{x_1}),P({y_4}_{x_2}),\\
&&UB(P({y_4}_{x_2},x_1,y_1))+UB(P({y_1}_{x_1},x_2,y_4))\}\\
&=&\min\{205/600,364/600,0.005+0.034\}\\
&=&0.039.
\end{eqnarray*}
Thus,
\begin{eqnarray*}
0\le P({y_1}_{x_1},{y_4}_{x_2})\le 0.039.
\end{eqnarray*}
Also by Theorem \ref{thm8},
\begin{eqnarray*}
&&P({y_2}_{x_1},{y_4}_{x_2})\\
&\ge&\max\{0, P({y_2}_{x_1})+P({y_4}_{x_2})-1,\\
&&LB(P({y_4}_{x_2},x_1,y_2))+LB(P({y_2}_{x_1},x_2,y_4))\}\\
&=&\max\{0,46/600+364/600-1,0.037+0\}\\
&=&0.037.
\end{eqnarray*}
and,
\begin{eqnarray*}
&&P({y_2}_{x_1},{y_4}_{x_2})\\
&\le&\min\{P({y_2}_{x_1}),P({y_4}_{x_2}),\\
&&UB(P({y_4}_{x_2},x_1,y_2))+UB(P({y_2}_{x_1},x_2,y_4))\}\\
&=&\min\{46/600,364/600,0.062+0.015\}\\
&=&0.077.
\end{eqnarray*}
Thus,
\begin{eqnarray*}
0.037\le P({y_2}_{x_1},{y_4}_{x_2})\le 0.077.
\end{eqnarray*}
Again by Theorem \ref{thm8},
\begin{eqnarray*}
&&P({y_3}_{x_1},{y_4}_{x_2})\\
&\ge&\max\{0, P({y_3}_{x_1})+P({y_4}_{x_2})-1,\\
&&LB(P({y_4}_{x_2},x_1,y_3))+LB(P({y_3}_{x_1},x_2,y_4))\}\\
&=&\max\{0,343/600+364/600-1,0.502+0\}\\
&=&0.502.
\end{eqnarray*}
and,
\begin{eqnarray*}
&&P({y_3}_{x_1},{y_4}_{x_2})\\
&\le&\min\{P({y_3}_{x_1}),P({y_4}_{x_2}),\\
&&UB(P({y_4}_{x_2},x_1,y_3))+UB(P({y_3}_{x_1},x_2,y_4))\}\\
&=&\min\{343/600,364/600,0.527+0.034\}\\
&=&0.561.
\end{eqnarray*}
Thus,
\begin{eqnarray*}
0.502\le P({y_3}_{x_1},{y_4}_{x_2})\le 0.561.
\end{eqnarray*}

\subsection{Distribution Generating Algorithm}
Here, the sample distribution generating algorithm in the simulated study is presented. It generated both experimental and observational data compatible with the fractions of response types of individuals. The data satisfy the general relation between experimental and observational data.
\begin{algorithm}[tb]
\caption{Generate samples for simulated study}
\label{alg1}
\textbf{Input}: $num$, number of samples needed.\\
\textbf{Output}: $num$ sample distributions (observational data and experimental data).
\begin{algorithmic}[1] %[1] enables line numbers
\STATE $count = 0$;
\WHILE {$count<num$}
    \STATE //$rand(0,1)$ is the function that random uniformly generate a number from $0$ to $1$.
    \STATE $a = []$;
    \FOR{$i=1$ to $8$}
        \STATE $a.append(rand(0,1))$;
    \ENDFOR    
    \STATE $a.append(1.0)$;
    \STATE $a.sort()$;
    \STATE //$f$ is the fractions of response types of individuals, $f[0]=P({y_1}_{x_1},{y_1}_{x_2}),...,f[8]=P({y_3}_{x_1},{y_3}_{x_2}).$
    \STATE $f = []$;
    \STATE $f[0] = a[0]$;
    \FOR{$i=1$ to $8$}
        \STATE $f[i] = a[i] - a[i-1]$;
    \ENDFOR
    \STATE // Generate experimental data.
    \STATE $P({y_1}_{x_1})=f[0]+f[1]+f[2]$;
    \STATE $P({y_2}_{x_1})=f[3]+f[4]+f[5]$;
    \STATE $P({y_3}_{x_1})=f[6]+f[7]+f[8]$;
    \STATE $P({y_1}_{x_2})=f[0]+f[3]+f[6]$;
    \STATE $P({y_2}_{x_2})=f[1]+f[4]+f[7]$;
    \STATE $P({y_3}_{x_2})=f[2]+f[5]+f[8]$;
    \STATE // Generate observational data.
    \STATE $P({x_1},{y_1})=rand(0,P({y_1}_{x_1}))$;
    \STATE $P({x_1},{y_2})=rand(0,P({y_2}_{x_1}))$;
    \STATE $P(x_1)=rand(P({x_1},{y_1})+P({x_1},{y_2}),\min\{P({x_1},{y_1})+1-P({y_1}_{x_1}),P({x_1},{y_2})+1-P({y_1}_{x_2})\})$;
    \STATE $P({x_1},{y_3})=P(x_1)-P({x_1},{y_1})-P({x_1},{y_2})$;
    \STATE $P(x_2) = 1-P(x_1)$
    \STATE $P({x_2},{y_1})=rand(0,\min\{P({y_1}_{x_2}),P(x_2)\})$;
    \STATE $P({x_2},{y_2})=rand(0,\min\{P({y_2}_{x_2}),P(x_2)-P({x_2},{y_1})\})$;
    \STATE $P({x_2},{y_3})=P(x_2)-P({x_2},{y_1})-P({x_2},{y_2})$;
    \STATE //Validate the data, the experimental data and observational data should satisfies the following: $P(x,y)\le P(y_x)\le P(x,y)+1-P(x)$.
    \STATE $mark = True$
    \FOR{$i=1$ to $3$}
        \FOR{$j=1$ to $2$}
            \IF {$P({y_i}_{x_j})<P(x_j,y_i)$ or $P({y_i}_{x_j})>P(x_j,y_i)+1-P(x_j)$}
                \STATE $mark = False$;
            \ENDIF
        \ENDFOR
    \ENDFOR
    \IF{$mark == False$}
        \STATE $continue$;
    \ENDIF
    \STATE $count = count + 1$;
    % \STATE // Each $c_i$ corresponding to a sample distribution.
    % \STATE // The following are experimental data that satisfied the general bounds provided by Tian and Pearl.
    % \STATE $P(y|do(x),z,c_i)=rand(0,1)\times \frac{t_2}{t_1+t_2}+\frac{o_1}{t_1+t_2}$;
    % \STATE $P(y|do(x'),z,c_i)=rand(0,1)\times \frac{t_1}{t_1+t_2}+\frac{o_2}{t_1+t_2}$;
    % \STATE $P(y|do(x),z',c_i)=rand(0,1)\times \frac{t_4}{t_3+t_4}+\frac{o_3}{t_3+t_4}$;
    % \STATE $P(y|do(x'),z',c_i)=rand(0,1)\times \frac{t_3}{t_3+t_4}+\frac{o_4}{t_3+t_4}$;
    % \STATE // The following are observational data.
    % \STATE $P(x,y,z|c_i)=o_1/1000$;
    % \STATE $P(x,y,z'|c_i)=o_3/1000$;
    % \STATE $P(x,y',z|c_i)=(t_1-o_1)/1000$;
    % \STATE $P(x,y',z'|c_i)=(t_3-o_3)/1000$;
    % \STATE $P(x',y,z|c_i)=o_2/1000$;
    % \STATE $P(x',y,z'|c_i)=o_4/1000$;
    % \STATE $P(x',y',z|c_i)=(t_2-o_2)/1000$;
    % \STATE $P(x',y',z'|c_i)=(t_4-o_4)/1000$;
\ENDWHILE
\end{algorithmic}
\end{algorithm}
\end{document}